\documentclass[11pt,letterpaper]{article}

\usepackage[T1]{fontenc}
\usepackage[utf8]{inputenc}
\usepackage[margin=1in]{geometry}
\usepackage{times}
\usepackage{amsmath,amssymb,bm}
\usepackage{graphicx,booktabs,tabularx}
\usepackage[table]{xcolor}
\usepackage{xspace,microtype}
\usepackage[authoryear,round]{natbib}
\usepackage{xurl}
\usepackage{authblk}
\usepackage[hidelinks]{hyperref}
\newcommand{\secrref}[1]{Sec.~\ref{#1}\xspace}
\newcommand{\appref}[1]{Appendix~\ref{#1}\xspace}
\newcommand{\figrref}[1]{Fig.~\ref{#1}\xspace}
\newcommand{\heading}[1]{{\vspace{3pt}\noindent\textbf{#1}}}
\newcommand{\tabref}[1]{\mbox{Table~\ref{#1}}}
\newcommand{\equaref}[1]{Eq.~\ref{#1}\xspace}

\newenvironment{packeditemize}{%
  \begin{list}{$\bullet$}{%
    \setlength{\labelwidth}{4pt}%
    \setlength{\itemsep}{0pt}%
    \setlength{\leftmargin}{\labelwidth}%
    \addtolength{\leftmargin}{\labelsep}%
    \setlength{\parindent}{0pt}%
    \setlength{\listparindent}{\parindent}%
    \setlength{\parsep}{0pt}%
    \setlength{\topsep}{1pt}%
  }%
}{\end{list}}

\newcommand{\rep}{\ensuremath{\mathsf{MiniRep}}\xspace}
\newcommand{\nagents}{n\xspace}
\newcommand{\nfaulty}{f\xspace}
\newcommand{\ndebate}{R\xspace}
\newcommand{\proposal}{\ensuremath{v}\xspace}
\newcommand{\param}{\chi}
\newcommand{\task}{\ensuremath{\tau}\xspace}
\newcommand{\cont}{\ensuremath{c}\xspace}
\newcommand{\reason}{\ensuremath{\rho}\xspace}
\newcommand{\outspace}{\ensuremath{\mathcal{Y}}\xspace}
\newcommand{\state}{\ensuremath{\sigma}\xspace}
\newcommand{\coalition}{\ensuremath{B}\xspace}
\newcommand{\agents}{\mathcal{P}\xspace}
\newcommand{\adv}{\ensuremath{\mathcal{A}}\xspace}
\newcommand{\aggfunc}{\mathsf{Agg}\xspace}
\newcommand{\ctx}{\mathit{ctx}\xspace}
\newcommand{\propose}{\emph{Propose}\xspace}
\newcommand{\debate}{\emph{Debate}\xspace}
\newcommand{\decide}{\emph{Decide}\xspace}

\date{}

\title{\textbf{\rep: Robust Reputation-Based Aggregation\\for Multi-Agent Debate}}
\hypersetup{
  pdftitle={MiniRep: Robust Reputation-Based Aggregation for Multi-Agent Debate},
  pdfsubject={Reputation-based aggregation for multi-agent debate},
  pdfkeywords={multi-agent debate, reputation, robust aggregation}
}
\author[1]{Jiaming Zhang}
\author[1]{Yuwan Liu}
\author[1]{Yue Huang}
\author[1]{Sisi Duan}

\affil[1]{Tsinghua University}

\hypersetup{pdfauthor={First Author, Second Author, Third Author}}

\newcommand{\codeavailability}{}

\begin{document}
\maketitle

\begin{abstract}
Autonomous agents powered by large language models (LLMs) are rapidly evolving into an open agentic ecosystem. To support trustworthy collaboration, industry initiatives increasingly assess agent reputation from past behavior and provide performance leaderboards. However, reputation derived from past performance may not reliably predict an agent's behavior on new tasks, particularly when malicious agents can adapt their behavior and influence other agents during collaboration.

We study reputation in multi-agent debate (MAD), where multiple agents answer the same query,  debate to improve their answers, and aggregate them into a final output. We present \ensuremath{\mathsf{MiniRep}}, a reputation-based aggregation system for MAD under malicious agents. To ground our threat model in established research, we construct an attack taxonomy drawing on reputation-system attacks and software-testing mutation operators, covering strategic exploitation of reputation and subtle corruption of agent proposals. Guided by this taxonomy, \ensuremath{\mathsf{MiniRep}} evaluates agents based on both their behavior on the current task and their reputation over time, while preventing groups of agents with highly similar responses from dominating the final decision. We assess \ensuremath{\mathsf{MiniRep}} across diverse tasks, LLM-agent compositions, corruption placements, and attack types drawn from our taxonomy. Our experimental results show that, \ensuremath{\mathsf{MiniRep}} outperforms both conventional MAD aggregation and conventional reputation-based approaches on MATH no matter being attacked or not. Also, under a heterogeneous 10-agent setting on MATH, \ensuremath{\mathsf{MiniRep}} outperforms all baselines in all 28 attack conditions.

\end{abstract}

\section{Introduction} \label{sec:intro}
Autonomous agents~\citep{wang2023survey, chen2024internet} powered by large language models (LLMs) are evolving from standalone tools into an open agentic ecosystem. Gartner projects that the average Fortune 500 enterprise will operate more than 150,000 AI agents by 2028~\citep{lin2026too_many_agents}. 
 As agents are developed by different providers and increasingly interact across organizational boundaries, several industry initiatives have emerged to establish standards and infrastructure for agents to discover, connect, and interact with each other. Notable examples include the Agent2Agent (A2A) protocol~\citep{a2a_protocol} and the ERC-8004 standard~\citep{erc8004}, which enable cross-provider agent interoperability and provide infrastructure for agent discovery and trust, respectively.

Along with the emergence of such open ecosystems, establishing trust among agents becomes increasingly important. Several emerging platforms therefore provide mechanisms to assess the \textit{reputation} of LLM agents. For instance, ERC-8004 provides a Reputation Registry, a standard interface for posting and retrieving agent feedback.
Several platforms and ongoing projects compute reputation scores or trust metrics from agent behavior and feedback~\citep{credence, said, chishti2026agentreputation, agentrepsdk}. Agent marketplaces such as AWS Marketplace~\citep{aws_agent_marketplace} also allow customers to rate agent products, resembling reputation mechanisms in conventional online marketplaces.

Reputation is not a new concept. Reputation systems have long been studied and deployed in peer-to-peer networks~\citep{kamvar2003eigentrust}, e-commerce~\citep{resnick2002trust}, and software agents~\citep{huynh2006integrated,klos2003decentralized,mui2002notions}. A common principle is to estimate an entity's future reliability from its past behavior or feedback. For example, Beta Reputation~\citep{josang2002beta} models the probability of satisfactory behavior using a Beta distribution updated from positive and negative feedback. TrueSkill~\citep{herbrich2007trueskill} maintains a probabilistic skill rating for each player in online games and updates the ratings based on past competition outcomes. Emerging LLM agent reputation systems mentioned above largely adopt this history-based principle by deriving reputation from historical task outcomes, interactions, or feedback.

Thus, an interesting question is: \textit{Can reputation derived from past performance reliably guide collective decisions when LLM agents may adapt their behavior or coordinate their responses?} Indeed, unlike conventional reputation settings that often involve relatively well-defined interactions, LLM agents autonomously operate on diverse and open-ended tasks, making their reliability highly task-dependent. Moreover, their autonomy, memory, tool use, and interactions with other agents introduce new attack surfaces~\citep{zhang2025agent,agentdojo,kavathekar-etal-2026-tamas}. These differences become particularly important when reputation determines an agent's influence on collective decisions. 

In this paper, we focus on reputation at the aggregation stage of \textit{multi-agent debate} (MAD), rather than for general-purpose LLM agents. In MAD, multiple agents independently respond to the same query, debate to improve their results, and aggregate them into a final output~\citep{du2024multiagent,li2024more}. Building on earlier work on AI debate towards safe AI~\citep{irving2018ai}, the problem has evolved into a general class of methods for improving LLM reasoning and factuality across diverse tasks~\citep{du2024multiagent,kaesberg2025voting, khan2024debating,choi2025vote}. A unique feature in such a paradigm is that all the agents execute the \textit{same task}, even when powered by different underlying LLMs. This common evaluation context makes agent behaviors directly comparable within a shared task context, providing a natural setting for studying and exploiting agent reputation. 

We present \rep, a reputation-based aggregation system for MAD. We consider malicious agents that seek to manipulate MAD outputs away from the correct answers. Although some efforts have been made in identifying vulnerabilities of MAD, the area is relatively new, and there is no unified taxonomy of MAD vulnerabilities. We therefore construct our attack taxonomy from two complementary sources: established attacks on reputation systems~\citep{hoffman2009survey}, which capture strategic manipulation across agents and over time, and mutation operators from software testing~\citep{petrovic2018state,just2014mutants}, which capture subtle proposal-level corruptions. 
An interesting aspect is that for each attack type, we identify related mechanisms or agent behaviors reported in prior work, grounding the taxonomy in existing evidence rather than purely synthetic constructions. We summarize our taxonomy and related analogs in \tabref{tab:attack-overview}. Based on such a taxonomy, we then design a new way of building reputation scores for the agents, taking into consideration how agents can launch the attacks. \rep carefully selects a group of agents, the proposals of which are aggregated. Particular emphasis is placed on limiting the influence of agent groups with the same underlying LLM. Also, agents are rated based on both their reputation over time and behavior on the current task, such that temporal deviation can be caught. Thus, current evidence can reduce an agent's influence even after it has accumulated a strong reputation. 

We assess \rep on three datasets, GoEmotions~\citep{demszky2020goemotions}, MATH~\citep{hendrycks2021math}, and HumanEval Pro~\citep{yu2025humanevalpro}, under four LLM-agent compositions (i.e., using a combination of LLMs for the agents), four corruption placements (corrupting different agents), and seven attack types summarized in \tabref{tab:attack-overview}, yielding 112 attack conditions per dataset. We compare \rep with three MAD aggregation approaches (Uniform Majority~\citep{kaesberg2025voting}, Uniform Random-$k$, and Single-Metric), and four reputation systems (EigenTrust~\citep{kamvar2003eigentrust}, Beta~\citep{josang2002beta}, TrueSkill~\citep{herbrich2007trueskill}, and Babylon~\citep{babylon2026software}).  In these controlled aggregation runs, \rep has the highest observed average attacked score on three of the four task metrics, with mixed clean-task results. The most notable results are on MATH: \rep achieves 66.75\% accuracy in clean runs, 7.50 percentage points higher than the best baseline, and 61.95\% accuracy in attacked runs, compared with 54.37\% for the best baseline. Across the 112 MATH attack conditions, \rep obtains a higher score than every baseline in 84 conditions. Also, under a heterogeneous 10-agent setting on MATH, \ensuremath{\mathsf{MiniRep}} outperforms all baselines in all 28 attack conditions. 
\codeavailability

\begin{table}[t]
\centering
\caption{Attack taxonomy studied in this work. Reputation classes follow~\citet{hoffman2009survey}. The last column lists related mechanisms reported in prior work; these studies do not explicitly target agent reputation. $^\star$ On-off and Adaptive adapt self-promoting attacks to MAD.}
\label{tab:attack-overview}
\footnotesize
\setlength{\tabcolsep}{3.5pt}
\renewcommand{\arraystretch}{1.08}
\begin{tabularx}{\linewidth}{
>{\raggedright\arraybackslash}p{1.45cm}
>{\raggedright\arraybackslash}p{4.05cm}
>{\centering\arraybackslash}p{1.60cm}
>{\raggedright\arraybackslash}X}
\toprule
\textbf{Attack} & \textbf{Instantiation} & \textbf{Reputation class} & \textbf{Related analogue} \\
\midrule
Random coordinated & Malicious agents submit the same randomly selected incorrect answer. & Orchestrated & Adversarial peers can induce conformity and degrade agent decisions~\citep{ko2026social}; incorrect proposals can propagate during MAD~\citep{cui2026freemad}. \\
Optimized coordinated & Malicious agents search for an incorrect proposal that maximizes its estimated impact on the final output. & Orchestrated & Adversarial agents can generate candidate arguments and select the most persuasive one~\citep{kraidia2026collaboration}. \\
Diverse collusion & Malicious agents submit distinct proposals that support the same incorrect conclusion. & Orchestrated & Colluding agents can provide distinct contributions toward a common objective and induce false conclusions~\citep{hu2026lying}. \\
On-off & Malicious agents first build reputation and then attack on every subsequent task. & Self-promoting$^\star$ & Agents can first build peer trust and later inject fabricated information~\citep{park2026data}. \\
Adaptive & Malicious agents first build reputation. During an attack, only a selected subset deviates from the system specification. & Self-promoting$^\star$ & Manipulating one agent can be sufficient to mislead a multi-agent system~\citep{liu2025can}. \\
Arithmetic operator & A malicious agent changes one arithmetic operation in an honestly generated proposal. & N/A & Arithmetic and relational operator replacement can introduce small but consequential faults~\citep{petrovic2018state}. \\
Boundary value & A malicious agent modifies a boundary value or literal in an honestly generated proposal. & N/A & Literal replacement can introduce faults into otherwise unchanged programs~\citep{just2014mutants}. \\
\bottomrule
\end{tabularx}
\end{table}

\section{Problem Background and Threat Model} \label{sec:model}

\heading{Multi-agent debate (MAD).}
Typical MAD protocols organize each task into three phases:
\propose, in which agents independently produce candidate answers;
\debate, which, when performed, allows agents to exchange information and revise their responses over one or more rounds; 
and \decide, in which the resulting responses are aggregated into a final
output~\citep{du2024multiagent,chan2023chateval}.
We consider a sequence of tasks indexed by $t$. Let
$\agents=\{1,\ldots,\nagents\}$ denote the set of agents, and let
$\task_t=(\ctx_t,\param_t)$ denote task $t$, where $\ctx_t$ contains the \textit{public evidence}, i.e., information provided to all agents, and $\param_t$ contains task-specific parameters. In the \propose phase, each agent $i$ produces a response
$\proposal_{i,t}=(\cont_{i,t},\reason_{i,t})$, consisting of a payload $\cont_{i,t}\in\outspace$ and an optional explanation $\reason_{i,t}$, where $\reason_{i,t}=\varnothing$ if no explanation is provided. After
$\ndebate$ debate rounds, the aggregation rule $\aggfunc$ first determines a pool
$P_t\subseteq\agents$ of eligible agents and then maps their responses, the
task context, and any persistent state $\state_{t-1}$ to a final response. A
stateless aggregation rule simply omits $\state_{t-1}$. 

The \decide phase is often the most distinctive phase of MAD protocols. Common
approaches include equal-weight voting~\citep{wang2023selfconsistency,
kaesberg2025voting,li2024more}; model-based aggregation, in which an evaluator
such as an LLM judge assesses the candidate responses and produces or selects
the final answer~\citep{chan2023chateval,hu2026multiagent};
confidence-weighted voting, which gives greater influence to responses
associated with higher confidence~\citep{zheng2026rethinking}; and
history-based aggregation, which weights agents according to their performance on previous tasks~\citep{dehghankar2025credibility}. In this work, we focus on reputation-based aggregation in the \decide phase. We compare our approach
against three aggregation baselines. Uniform
Majority~\citep{kaesberg2025voting} aggregates the responses of all identities
using equal weights. Uniform Random-$k$ is a size-matched control that samples
$k$ identities independently of their past performance and aggregates their
responses using equal weights. Single-Metric maintains a single historical
quality score for each agent and uses these scores to rank and weight the agents. Thus, Single-Metric can also be viewed as a reputation-based approach based on past performance. We formally define these baselines in \appref{app:baseline}.

In our system model, each agent retains a persistent identity across tasks, allowing its reputation to be updated over time. We refer to agents instantiated from the same underlying model as a \emph{clone group}, and to the known partition of $\agents$ into such groups as the \emph{clone-group mapping}.

\heading{Assumptions and threat model.} The adversary controls a fixed set $\coalition\subseteq\agents$ of at most $\nfaulty$ agents within each run. On task $t$, it may activate any subset $\coalition_t\subseteq\coalition$.  Its goal is to manipulate the output of MAD, causing it to deviate from the output that would be produced if all agents behaved honestly. We assume that each malicious agent always participates in MAD and submits its candidate answer on time. A submitted proposal is observed consistently by the client and all agents. All agents also receive the same task context, including any public evidence. However, the candidate answer may deviate arbitrarily from the answer that the agent would produce if it behaved honestly. A malicious agent may behave honestly to accumulate reputation and begin attacking at any time chosen by $\adv$.
For every task, the system uses four categories of information: the proposals $\{\proposal_i\}$, the public evidence in $\ctx$, observable properties derived from the submitted proposals, and the clone-group mapping. The use of observable outcomes and behavior as reputation signals follows conventional reputation systems~\citep{mui2002notions,hoffman2009survey}.

Our assumptions are practical, namely that all malicious agents submit responses and that the clone-group mapping is available to the system. First, the assumption that malicious agents always respond is reasonable within our threat model: skipping a proposal cannot directly inject adversarial content into the aggregation~\citep{feldman2004robust,hoffman2009survey}.  
Second, assuming that the clone-group mapping is available is reasonable in a managed agent ecosystem, where an agent registry can expose registered identity, provider, or model information~\citep{erc8004}.

\begin{figure}[t]
    \centering
    \includegraphics[width=\linewidth]{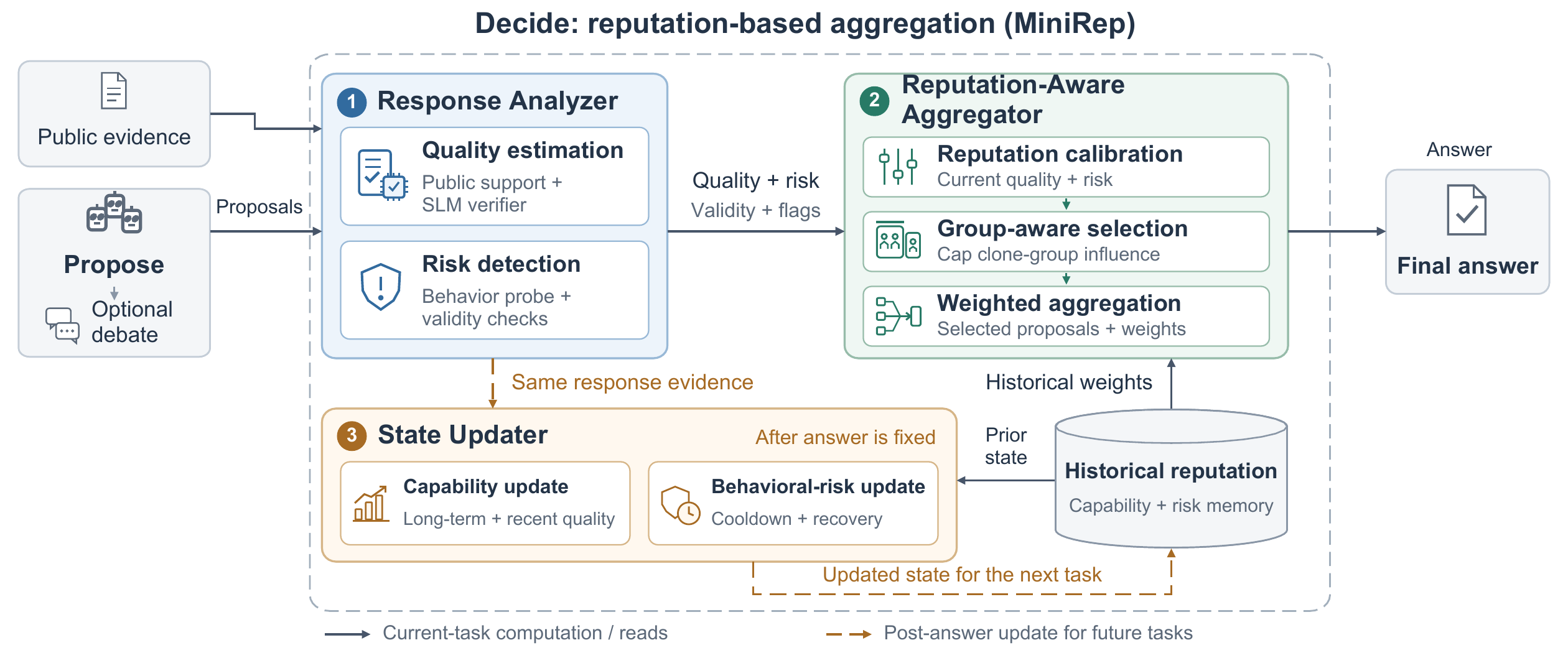}
    \caption{Overview of \rep.}
    \label{fig:overview}
    \vspace{-6pt}
\end{figure}

\section{Design of \rep} \label{sec:design}
\heading{Attack classes.} As mentioned in the introduction, we construct our attack taxonomy based on established attacks on reputation systems~\citep{hoffman2009survey} and mutation operators from software testing~\citep{petrovic2018state,just2014mutants}, as summarized in \tabref{tab:attack-overview}.  
These two sources capture complementary aspects of the threat model. First, because reputation determines an agent's influence on aggregation, adversaries may manipulate reputation to increase the impact of their proposals. Second, mutation operators provide systematic ways to introduce localized but consequential errors that may be difficult to detect in otherwise plausible proposals.  
Accordingly, we organize the attacks into three groups. First, random coordinated, optimized coordinated, and diverse collusion vary how malicious agents construct and coordinate their proposals within a task. Second, on-off and adaptive attacks manipulate reputation over time to increase adversarial influence. Finally, arithmetic-operator and boundary-value attacks apply simple mutation operators to corrupt otherwise honestly generated proposals.

\heading{Overview.} 
We show in \figrref{fig:overview} the workflow of \rep.
Recall that we use a reputation-based aggregation for the \decide phase only. Thus, we can simply follow any existing approach for the \propose and \debate phases. \

Suppose that we already have a reputation system that scores each agent's capability. Let $w^{hist}_{i,t}$ denote the reputation-derived
weight of agent $i$, and let
\(
\mathbf{w}^{hist}_t
=
\left(
w^{hist}_{1,t},
\ldots,
w^{hist}_{\nagents,t}
\right)
\)
be the weights of all agents in $\agents$.
A naive reputation-based approach directly uses these historical weights to
aggregate the agents' current proposals as 
\begin{equation}
\hat y_t^{\mathrm{naive}}
=
\aggfunc\!\left(
\{\proposal_{i,t}\}_{i\in\agents},
\ctx_t;
\mathbf{w}^{hist}_t
\right).
\label{eq:naive-reputation-aggregation}
\end{equation}
When conventional reputation systems rate agents based only on past performance, both self-promoting and orchestrated attacks can manipulate the aggregation output. For instance, a malicious agent with a powerful underlying LLM can stay honest to accumulate a high reputation, and then inject a wrong answer into a subsequent task to directly manipulate the output. 

\heading{A running example.}
Consider a level-4 MATH problem asking for the center of $x^2-6x+y^2+2y=9$~\citep{hendrycks2021math}. Completing the square gives
$(x-3)^2+(y+1)^2=19$, so the answer is $(3,-1)$. In a 10-agent debate, suppose that three high-reputation but malicious agents submit an incorrect answer $(3,1)$ while the other seven submit $(3,-1)$. The naive weighted vote outputs $(3,1)$ if the three malicious agents carry a higher accumulated historical weight than that of the seven honest agents.

\heading{Our approach.} In our work, we retain this basic idea in a \textit{Reputation-Aware Aggregator} module, but additionally include a \textit{Response Analyzer} that evaluates the quality of each current proposal and whether it shows signs of malicious behavior before aggregation. After the Reputation-Aware Aggregator produces the answer $\hat y_t$, the \textit{State Updater} uses the signals produced by the Response Analyzer to update each agent's historical quality and behavioral-risk (i.e., previously detected suspicious behavior) records. These records determine the agents' historical weights on subsequent tasks. The three modules have distinct roles: the Response Analyzer assesses current proposals, the Reputation-Aware Aggregator adjusts their weights and limits clone-group influence, and the State Updater updates the historical reputation.

More formally, the three modules work as follows. We give the design details in \appref{app:design}. 
\begin{packeditemize}

\item[\textbf{Response Analyzer.}] The Response Analyzer takes as input the proposals ${\proposal_{i,t}}_{i\in\agents}$, the public evidence $\ctx_t$, and the clone-group mapping, and outputs quality, risk, validity, and blocking signals.  Here, the public evidence refers to the information provided to all agents as part of the task, such as a question statement, a set of given facts, or a function specification.

The analyzer first evaluates each proposal. It assigns the proposal a quality score $q_{i,t}$ based on its agreement with the other proposals and an independent assessment of its semantic content. We compute a public-support score from the current proposals without using the ground-truth answer or hidden tests. This score combines agreement among proposals with the overall answer distribution. Also, we group proposals by clone groups so that proposals by agents from the same clone group are not counted as independent sources of evidence. Finally, to check the semantic content, we use a semantic verifier instantiated by a small language model (a 1.5B model in our experiments). The goal is to estimate whether the proposal is correct based on the task and the content of the proposal.

The analyzer also estimates whether the proposal exhibits malicious or manipulative behavior via a \textit{behavior model} implemented as a multilayer perceptron (MLP)~\citep{rumelhart1986learning}. The behavior model checks the current proposal for signs such as instruction override and, when debate revisions are available, unexpected changes across rounds. 

A proposal is blocked on the current task if no candidate answer can be extracted, or if the shared detector flags it using combined correctness and behavior signals. We show the exact rule in \appref{app:design:fusion}.

\item[\textbf{Reputation-Aware Aggregator.}] The Reputation-Aware Aggregator takes the quality, risk, validity, and blocking signals produced by the Response Analyzer, together with the historical weights ${w^{hist}_{i,t}}_{i\in\agents}$ and the clone-group mapping, and outputs an aggregated result.

For each agent, the aggregator adjusts its historical weight using the proposal's current quality relative to the other valid proposals, its current risk evidence, and the agent's recorded behavioral risk.  Blocked proposals, recently suspicious agents, and responses without an extractable answer receive additional weight penalties. Thus, a high historical weight alone does not guarantee high influence on the current task. The exact weighting and clone-group rules are deferred to \appref{app:design:weighting}.

After that, only $k$ out of $\nagents$ agents are selected into a pool $P_t\subseteq\agents$. Before selection, the aggregator limits the total influence of each clone group so that multiple agents instantiated from the same underlying model are not treated as independent sources of reputation. It then uses the resulting weights to select the pool and determine the final aggregation weights $\{w_{i,t}\}_{i\in P_t}$. Based on the pool and the weights generated by the aggregator, the selected proposals are combined using reputation-weighted aggregation:
\begin{equation}
\hat y_t
=
\aggfunc\!\left(
\{\proposal_{i,t}\}_{i\in P_t},
\ctx_t;
\{w_{i,t}\}_{i\in P_t}
\right).
\label{eq:overview-rep-aggregation}
\end{equation}
The task-dependent implementations of $\aggfunc$ are given under ``Task-dependent aggregation rules'' in \appref{app:baseline}.
Recall that the naive reputation-based approach is recovered by selecting all agents, $P_t=\agents$, and directly using their historical weights, $w_{i,t}=w^{hist}_{i,t}$.

\item[\textbf{State Updater.}] The State Updater runs after the Reputation-Aware Aggregator produces $\hat y_t$. It uses the same quality, risk, validity, and blocking evidence produced by the Response Analyzer to update each agent's historical reputation for subsequent tasks.

For each agent, the updater separately records its long-term quality, recent quality, and previously detected suspicious behavior. Briefly speaking, the long-term quality and recent quality are computed based on the quality scores $q_{i,t}$ of the agent's valid proposals across previous tasks. Consistently high-quality proposals gradually improve the agent's reputation, whereas suspicious behavior reduces its influence on future tasks. Also, an unavailable or not well-formed response does not provide a new quality or risk observation.

The updated records form $\state_t$ and determine the historical weight $w^{hist}_{i,t+1}$ used on the next task. Note that $\state_t$ is produced only after $\hat y_t$ is fixed, so it cannot affect the answer to task $t$.
\end{packeditemize}

\section{Experiments}
\label{sec:evaluation}

\subsection{Experimental Setup}
\label{sec:evaluation:setup}

\begin{table}[!htb]
\centering
\caption{Experimental setup. Attack abbreviations follow the order in \tabref{tab:attack-overview}.}
\label{tab:design}
\footnotesize
\setlength{\tabcolsep}{5pt}
\begin{tabularx}{\linewidth}{@{}>{\raggedright\arraybackslash}p{0.22\linewidth}>{\raggedright\arraybackslash}X@{}}
\toprule
\textbf{Name (\#)} & \textbf{Settings} \\
\midrule
Dataset (3) & GoEmotions~\citep{demszky2020goemotions}, MATH (levels 4--5)~\citep{hendrycks2021math}, HumanEval Pro~\citep{yu2025humanevalpro} \\
\addlinespace[2pt]
Composition (4) & all strong, 7-strong$+$3-weak (7S+3W), all weak, random \\
\addlinespace[2pt]
Placement (4) & B (top-ranked), W (bottom-ranked), R1, R2 (random) \\
\addlinespace[2pt]
Attacks (7 + clean) & \textsc{Rand}~\citep{hoffman2009survey,ko2026social}, \textsc{Strong}~\citep{kraidia2026collaboration}, \textsc{Div}~\citep{hu2026lying}, \textsc{OnOff}~\citep{park2026data}, \textsc{Adapt}~\citep{hoffman2009survey,liu2025can}, \textsc{Arith}~\citep{petrovic2018state}, \textsc{Bound}~\citep{just2014mutants} \\
\bottomrule
\end{tabularx}
\end{table}

\heading{Benchmarks.} We use three datasets: GoEmotions~\citep{demszky2020goemotions} for fine-grained multi-label emotion classification, MATH~\citep{hendrycks2021math} for competition-level mathematics (levels 4--5), and HumanEval Pro~\citep{yu2025humanevalpro} for Python programming. For each dataset, we use the same fixed set of 100 tasks throughout. We provide examples from these datasets in \appref{app:experiment}.

\heading{Agent pool and composition.} We construct a pool of agents instantiated from seven API-accessible LLMs. Our goal is to evaluate reputation-based aggregation under meaningful differences in agent capability. We therefore select LLMs with a range of standalone scores on our datasets. This setting provides both correct and incorrect proposals for evaluating whether reputation-based aggregation assigns influence appropriately. 

We rank the models from strong to weak using standalone evaluations on each dataset. The ranking is computed separately for each dataset because the relative performance of the models varies across tasks. In our experiments, we assess four compositions according to these rankings as summarized in \tabref{tab:design}. We provide the model rankings and the construction of each ten-agent pool in \appref{app:experiment}.

\heading{Experimental setup and corruption placement.} For all experiments, we fix $\nagents=10$, $\nfaulty=3$, and $k=7$. We use four corruption placements: B and W compromise agents from higher- and lower-ranked models, respectively, while R1 and R2 are two random assignments. 

\heading{Baselines.} We assess seven baselines: three MAD aggregation baselines and four reputation systems. The MAD baselines are Uniform Majority (UMaj)~\citep{kaesberg2025voting}, Uniform Random-$k$ (URand), and Single-Metric (Single). Uniform Random-$k$ and Single-Metric are in-study controls for pool size and simple historical feedback, respectively. The reputation baselines are EigenTrust (Eigen)~\citep{kamvar2003eigentrust}, Beta Reputation (Beta)~\citep{josang2002beta}, TrueSkill (TS)~\citep{herbrich2007trueskill}, and Babylon~\citep{babylon2026software}. All stateful methods share post-task quality feedback; \rep also uses current-task signals before aggregation. We provide their definitions and adaptations in \appref{app:baseline}.

All methods replay the same cached proposals with $\ndebate=0$, isolating aggregation from within-task debate. Reputation still evolves across tasks, including the OnOff and Adapt schedules. We compare complete aggregation procedures, including current-task screening and the GoEmotions readout (\appref{app:baseline}), rather than individual reputation components. Additional analyses of held-out attacks and agent compositions appear in \appref{app:feedback-generalization} and \appref{app:composition-averages}. 

\heading{Metrics.} For GoEmotions~\citep{demszky2020goemotions}, we report exact-match accuracy and sample F1~\citep{sokolova2009systematic}. Exact-match accuracy requires the predicted and correct label sets to be identical. Sample F1 gives partial credit for overlapping labels. We use mathematical-equivalence accuracy for MATH~\citep{hendrycks2021math} and Pass@1~\citep{chen2021evaluating} for HumanEval Pro~\citep{yu2025humanevalpro}.

For each experiment, we compare a \textit{clean} run and an \textit{attacked} run, where the attack is launched only in the attacked run. With four corruption placements and seven attacks, we assess 28 attack conditions for each agent composition and 112 conditions per dataset. We define all the metrics we use in \appref{app:metric}.  %

\subsection{Evaluation Results}
\label{sec:evaluation:results}
We report descriptive point estimates; strict wins denote higher observed scores, not statistical significance. Additional results are given in \appref{sec:additional-results}. 

\begin{packeditemize}
    \item \rep can improve performance even in clean runs for some tasks. The clearest gain is on MATH, where \rep exceeds the best baseline by 7.50 percentage points in accuracy. This is consistent with selecting and weighting proposals using both historical performance and current quality, which allows \rep to downweight temporarily poor responses.
    \item The most consistent attacked-score advantage is on MATH. \rep achieves the highest average attacked score on three of the four task metrics and obtains 84 strict wins across the 112 MATH attack conditions. 
    \item  Even under self-promoting attacks, i.e., OnOff and Adapt in our evaluation, where the adversary first builds reputation and then attacks, \rep outperforms the best baseline in five of the eight evaluated conditions and ties it in two.
\end{packeditemize}

\heading{Overall performance of all experiments.} \tabref{tab:average_utility} summarizes average scores over the clean runs and 112 attack conditions per dataset. In clean runs, the largest gain is on MATH: \rep achieves 66.75\% accuracy, 7.50 percentage points above the best baseline. Clean GoEmotions F1 and HumanEval Pro Pass@1 are lower than the best baselines by 2.00 and 1.75 percentage points, respectively. Under attack, \rep has the highest observed GoEmotions accuracy (21.16\%), MATH accuracy (61.95\%), and HumanEval Pro Pass@1 (78.15\%), with only a 0.23-point edge on HumanEval Pro.

In \tabref{tab:main_result_summary}(a), we fix each agent composition and summarize the number of \textit{strict wins} among the 28 attack conditions. Our results show that \rep achieves broad performance gains across different agent compositions, with the most consistent improvements on MATH. Specifically, it strictly outperforms all baselines in all 28 attack conditions under the 7S+3W composition and in 27 of the 28 all-strong conditions. The gains also generally apply to other tasks: \rep wins 25 of the 28 7S+3W conditions for GoEmotions exact-match accuracy, 17 of the 28 all-weak conditions for GoEmotions F1, and 12 of the 28 7S+3W conditions for HumanEval Pro. These results are consistent with the Reputation-Aware Aggregator using the Response Analyzer's current-quality and risk signals to adjust historical weights. 

\begin{table}[!htb]
\centering
\caption{Average task scores (\%). The best evaluated baseline is selected separately for results in the clean and attacked runs. Bold marks the better score; full eight-method results are in \tabref{tab:average_utility_full}.}
\label{tab:average_utility}
\footnotesize
\setlength{\tabcolsep}{5pt}
\begin{tabularx}{\linewidth}{@{}>{\raggedright\arraybackslash}Xrrrr@{}}
\toprule
& \multicolumn{2}{c}{Clean} & \multicolumn{2}{c}{Attacked} \\
\cmidrule(lr){2-3}\cmidrule(l){4-5}
Dataset / metric & Best baseline & \rep & Best baseline & \rep \\
\midrule
GoEmotions / accuracy & 21.50 (Beta) & \textbf{21.75} & 19.59 (Beta) & \textbf{21.16} \\
GoEmotions / sample F1 & \textbf{36.42} (UMaj) & 34.42 & \textbf{34.87} (Beta) & 33.94 \\
MATH / accuracy & 59.25 (UMaj) & \textbf{66.75} & 54.37 (Beta) & \textbf{61.95} \\
HumanEval Pro / Pass@1 & \textbf{79.50} (UMaj) & 77.75 & 77.92 (Eigen) & \textbf{78.15} \\
\bottomrule
\end{tabularx}
\end{table}

\begin{table*}[t]
\centering
\caption{Performance summary against the best evaluated baseline $b$. Panel (a) reports strict wins out of 28 attack conditions for each composition. Panels (b)--(c) report attacked task scores (\%) where bold numbers denote the better value. The complete data are included in Tables~\ref{tab:configuration_wins}--\ref{tab:late_attack_code}, where repeated cells are highlighted.}
\label{tab:main_result_summary}
\footnotesize

\begin{minipage}[t]{0.50\textwidth}
\vspace{0pt}
\centering
\textbf{(a) Strict wins under different agent compositions.}\par
\vspace{2pt}
\setlength{\tabcolsep}{2.5pt}
\begin{tabularx}{\linewidth}{@{}>{\raggedright\arraybackslash}Xrr@{}}
\toprule
Setting & Best baseline $b$ & \rep \\
\midrule
GoEmotions Acc./7S+3W & 0/28 (all) & \textbf{25/28} \\
GoEmotions F1 / All weak & 8/28 (Single) & \textbf{17/28} \\
MATH / All strong & 0/28 (all) & \textbf{27/28} \\
MATH / 7S+3W & 0/28 (all) & \textbf{28/28} \\
HumanEval Pro / 7S+3W & 10/28 (Eigen) & \textbf{12/28} \\
\bottomrule
\end{tabularx}

\vspace{1pt}
{\scriptsize Strict task-score wins; ties are excluded.}
\end{minipage}
\hfill
\begin{minipage}[t]{0.485\textwidth}
\vspace{0pt}
\centering
\textbf{(b) Performance under the 7S+3W composition.}\par

\vspace{2pt}
\setlength{\tabcolsep}{2.5pt}
\begin{tabularx}{\linewidth}{@{}>{\raggedright\arraybackslash}Xrr@{}}
\toprule
Setting & Best baseline $b$ & \rep \\
\midrule
GoEmotions / B / Strong & \textbf{35.83} (Beta) & 33.00 \\
GoEmotions / B / Arith & 34.00 (Beta) & \textbf{34.90} \\
GoEmotions / B / Bound & 33.73 (Single) & \textbf{35.17} \\
MATH / R2 / Div & 55.00 (Babylon) & \textbf{73.00} \\
MATH / R2 / Bound & 54.00 (Babylon) & \textbf{69.00} \\
\bottomrule
\end{tabularx}

\vspace{1pt}
{\scriptsize Attacked F1 for GoEmotions and accuracy for MATH (\%).}
\end{minipage}

\vspace{4pt}
\begin{minipage}[t]{\textwidth}
\vspace{0pt}
\centering
\textbf{(c) Performance under OnOff and Adapt attacks on HumanEval Pro under the 7S+3W composition.}\par

\vspace{2pt}
\footnotesize
\setlength{\tabcolsep}{3pt}
\begin{tabularx}{\linewidth}{@{}>{\raggedright\arraybackslash}p{0.11\linewidth}*{8}{>{\centering\arraybackslash}X}@{}}
\toprule
& \shortstack{B/OnOff}
& \shortstack{B/Adapt}
& \shortstack{R1/OnOff}
& \shortstack{R1/Adapt}
& \shortstack{R2/OnOff}
& \shortstack{R2/Adapt}
& \shortstack{W/OnOff}
& \shortstack{W/Adapt} \\
\midrule
Baseline $b$
& Eigen & UMaj & Eigen & Eigen & Eigen & Eigen & Babylon & Babylon \\
Best $b$
& 78 & \textbf{81} & 81 & \textbf{82} & 79 & 80 & 80 & \textbf{81} \\
\rep
& \textbf{81} & \textbf{81} & \textbf{82} & 79 & \textbf{82} & \textbf{83} & \textbf{81} & \textbf{81} \\
\bottomrule
\end{tabularx}

\vspace{1pt}
{\scriptsize Attacked Pass@1 (\%).}
\end{minipage}

\vspace{2pt}
\begin{minipage}{0.98\textwidth}
\scriptsize
 In (a), $b$ has the largest baseline win count; in (b)--(c), it has the highest score for that condition. B, R1, R2, and W denote the four corruption placements. Attack abbreviations are listed in \tabref{tab:design}.
\end{minipage}
\end{table*}

\heading{Performance when capable agents are compromised.} We select five attack conditions in which the compromised agents include highly ranked models, all using the 7S+3W composition. For placement, we use B for GoEmotions and R2 for MATH. We report their attacked task scores in \tabref{tab:main_result_summary}(b). Compared with the best baseline, \rep achieves the highest score in four of the five conditions. In particular, it exceeds the strongest baseline by 18 and 15 percentage points under the MATH Div and Bound attacks, respectively. This result is consistent with our design: the Response Analyzer checks the current proposal, and the Reputation-Aware Aggregator adjusts its weight regardless of the agent's past reputation. The only exception is GoEmotions/Strong, where \rep scores 33.00\% F1 and Beta scores 35.83\%. This difference largely reflects their clean performance: Beta achieves 38.27\% F1 and \rep achieves 35.00\%. However, under attack, \rep has a smaller performance loss than Beta.

\heading{Performance under OnOff and Adapt attacks.} We evaluate the OnOff and Adapt attacks under all four corruption placements using the 7S+3W composition on HumanEval Pro, i.e.,  eight conditions in total. \tabref{tab:main_result_summary}(c) reports Pass@1 for all eight conditions. Our results show that \rep outperforms the best baseline in five out of eight conditions and ties it in two of them. The gains are 1--3 percentage points over baselines that already achieve 78\%--82\% Pass@1. The wins span all four corruption placements. This pattern is consistent with checking current proposals before aggregation and carrying detected risk forward to later tasks.

\section{Related Work}
\label{sec:related}
\heading{Attacks on agents and MAD.}
A growing body of work documents attacks on LLM agents: indirect prompt injection hijacks an agent through the data it reads \citep{greshake2023not, agentdojo}, backdoors are implanted into its memory or knowledge bases \citep{yang2024watch, chen2024agentpoison}, and social-engineering exploits propagate through agent interaction \citep{shapira2026agents, deng2025agents}.
Multi-agent debate inherits all of these channels and adds its own: agents can be driven to conform to a wrong majority \citep{amayuelas2024multiagent}.
Some defenses assess agents within a single task. Confidence-weighted consensus probes answer confidence from prompts and hidden states to tolerate Byzantine majorities in one instance \citep{zheng2026rethinking}. Blockchain-based coordination couples bookkeeping with multi-metric answer evaluation for one coordination run \citep{chen2024blockagents}.
Credibility scoring also learns agent weights from past query-answering contributions \citep{dehghankar2025credibility}.
\rep combines current-proposal quality and risk assessment with persistent reputation and clone-group control.

\heading{Classical reputation does not transfer.} Reputation has been studied for over two decades: electronic marketplaces aggregate transaction feedback into seller scores \citep{resnick2002trust}, peer-to-peer systems propagate trust transitively over interaction histories \citep{kamvar2003eigentrust}, probabilistic models ground reputation in statistics \citep{josang2002beta, herbrich2007trueskill}, and the software-agents community systematized these notions into integrated trust and reputation models \citep{mui2002notions, huynh2006integrated}.
Attacks and defenses are equally well understood: taxonomies classify attacks by the system component they target \citep{hoffman2009survey, koutrouli2012taxonomy}, cheap identities enable Sybil attacks in the absence of a central authority \citep{douceur2002sybil}, and TrustGuard counters strategic oscillation with PID-style update dynamics \citep{srivatsa2005trustguard}.
These results, however, do not transfer to multi-agent debate, for two reasons.
First, the feedback itself is missing: classical systems rate the observable outcome of a bilateral interaction, while in a debate the only visible outcome is the aggregated answer, so the per-agent contribution on which any reputation score must be built is never observed.
Second, the attack surface differs: classical defenses assume that honest behavior eventually shows up as good outcomes, whereas debate attacks such as conformity pressure and fabricated consensus produce outcomes that look reasonable while being wrong, so the outcome signal alone cannot separate honest agents from adversarial ones.
Recent reputation schemes for AI agents inherit the same outcome-based assumption \citep{ren2025reputation, lou2025drf, chishti2026agentreputation}, and deployed registries record feedback that is rarely grounded in verifiable interactions \citep{erc8004, xiong2026trustless}.
\rep therefore borrows only two classical ingredients: a prior-based statistical update in the style of \citet{josang2002beta} and fast reaction to recent negative behavior in the style of \citet{srivatsa2005trustguard}. It rebuilds the feedback layer itself, computing multi-signal behavioral evidence within each round and acting on it before aggregation.

\section{Conclusions}
We present \rep, a reputation-based aggregation approach for multi-agent debate (MAD). Our approach combines the quality of each agent's proposal on the current task with its historical performance. Across the evaluated aggregation settings, \rep achieves the highest average MATH accuracy in both clean and attacked runs. Results on the other tasks are mixed, including lower clean GoEmotions F1 and HumanEval Pro Pass@1 than the best baselines.

\clearpage
\bibliographystyle{plainnat}
\bibliography{references}

@misc{a2a_protocol,
  author = {{The Linux Foundation}},
  title = {Agent2Agent (A2A) Protocol},
  year = {2026},
  howpublished = {\url{https://a2a-protocol.org/latest/}},
  note = {Accessed: 2026-09-25.}
}

@inproceedings{agentdojo,
  author = {Debenedetti, Edoardo and Zhang, Jie and Balunovic, Mislav and Beurer-Kellner, Luca and Fischer, Marc and Tram\`{e}r, Florian},
  booktitle = {Advances in Neural Information Processing Systems},
  doi = {10.52202/079017-2636},
  editor = {A. Globerson and L. Mackey and D. Belgrave and A. Fan and U. Paquet and J. Tomczak and C. Zhang},
  pages = {82895--82920},
  publisher = {Curran Associates, Inc.},
  title = {AgentDojo: A Dynamic Environment to Evaluate Prompt Injection Attacks and Defenses for LLM Agents},
  url = {https://proceedings.neurips.cc/paper_files/paper/2024/hash/97091a5177d8dc64b1da8bf3e1f6fb54-Abstract-Datasets_and_Benchmarks_Track.html},
  volume = {37},
  year = {2024}
}

@misc{agentrepsdk,
  title = {agent-reputation-sdk: {Ethereum} {SDK} extensions for {ERC-8004} trustless agents},
  author = {Choe, Hanjoon},
  howpublished = {\url{https://github.com/hanjoonchoe/agent-reputation-sdk}},
  year = {2025},
  note = {Software repository. Accessed: 2026-09-25.}
}

@inproceedings{amayuelas2024multiagent,
    title = "{M}ulti{A}gent Collaboration Attack: Investigating Adversarial Attacks in Large Language Model Collaborations via Debate",
    author = "Amayuelas, Alfonso  and
      Yang, Xianjun  and
      Antoniades, Antonis  and
      Hua, Wenyue  and
      Pan, Liangming  and
      Wang, William Yang",
    editor = "Al-Onaizan, Yaser  and
      Bansal, Mohit  and
      Chen, Yun-Nung",
    booktitle = "Findings of the Association for Computational Linguistics: EMNLP 2024",
    month = nov,
    year = "2024",
    address = "Miami, Florida, USA",
    publisher = "Association for Computational Linguistics",
    url = "https://aclanthology.org/2024.findings-emnlp.407/",
    doi = "10.18653/v1/2024.findings-emnlp.407",
    pages = "6929--6948"
}

@misc{aws_agent_marketplace,
  author = {{Amazon Web Services}},
  title = {AI Agents and Tools in AWS Marketplace},
  howpublished = {\url{https://aws.amazon.com/marketplace/solutions/ai-agents-and-tools}},
  note = {Accessed: 2026-09-25.},
  year = {2026}
}

@misc{babylon2026software,
  author={{Babylon contributors}},
  title={{Babylon}: Reputation Calculation Service},
  howpublished={Official software repository, file \texttt{reputation-calculation-service.ts}},
  year={2026},
  url={https://github.com/BabylonSocial/babylon/blob/9b18096e3e69e68055ea4fc8cb0a1a94f80c2cef/packages/engine/src/reputation/reputation-calculation-service.ts},
  note={Accessed 2026-09-19. The cited source defines the composite score; the task-feedback adapter is specified in this paper}
}

@inproceedings{chan2023chateval,
  author = {Chan, Chi-Min and Chen, Weize and Su, Yusheng and Yu, Jianxuan and Xue, Wei and Zhang, Shanghang and Fu, Jie and Liu, Zhiyuan},
  booktitle = {International Conference on Learning Representations},
  editor = {B. Kim and Y. Yue and S. Chaudhuri and K. Fragkiadaki and M. Khan and Y. Sun},
  pages = {9079--9093},
  title = {ChatEval: Towards Better LLM-based Evaluators through Multi-Agent Debate},
  url = {https://proceedings.iclr.cc/paper_files/paper/2024/hash/25cc3adf8c85f7c70989cb8a97a691a7-Abstract-Conference.html},
  volume = {2024},
  year = {2024}
}

@article{chen2021evaluating,
  title = {Evaluating large language models trained on code},
  author = {Chen, Mark and Tworek, Jerry and Jun, Heewoo and Yuan, Qiming and Pinto, Henrique Ponde De Oliveira and Kaplan, Jared and Edwards, Harri and Burda, Yuri and Joseph, Nicholas and Brockman, Greg and others},
  journal = {arXiv preprint arXiv:2107.03374},
  year = {2021},
  url = {https://arxiv.org/abs/2107.03374}
}

@inproceedings{chen2024agentpoison,
  author = {Chen, Zhaorun and Xiang, Zhen and Xiao, Chaowei and Song, Dawn and Li, Bo},
  booktitle = {Advances in Neural Information Processing Systems},
  doi = {10.52202/079017-4136},
  editor = {A. Globerson and L. Mackey and D. Belgrave and A. Fan and U. Paquet and J. Tomczak and C. Zhang},
  pages = {130185--130213},
  publisher = {Curran Associates, Inc.},
  title = {AgentPoison: Red-teaming LLM Agents via Poisoning Memory or Knowledge Bases},
  url = {https://proceedings.neurips.cc/paper_files/paper/2024/hash/eb113910e9c3f6242541c1652e30dfd6-Abstract-Conference.html},
  volume = {37},
  year = {2024}
}

@inproceedings{chen2024blockagents,
author = {Chen, Bei and Li, Gaolei and Lin, Xi and Wang, Zheng and Li, Jianhua},
title = {BlockAgents: Towards Byzantine-Robust LLM-Based Multi-Agent Coordination via Blockchain},
year = {2024},
isbn = {9798400710117},
publisher = {Association for Computing Machinery},
address = {New York, NY, USA},
url = {https://doi.org/10.1145/3674399.3674445},
doi = {10.1145/3674399.3674445},
booktitle = {Proceedings of the ACM Turing Award Celebration Conference - China 2024},
pages = {187–192},
numpages = {6},
location = {Changsha, China},
series = {ACM-TURC '24}
}

@inproceedings{chen2024internet,
  author = {Chen, Weize and You, Ziming and Li, Ran and Guan, Yitong and Qian, Chen and Zhao, Chenyang and Yang, Cheng and Xie, Ruobing and Liu, Zhiyuan and Sun, Maosong},
  booktitle = {International Conference on Learning Representations},
  editor = {Y. Yue and A. Garg and N. Peng and F. Sha and R. Yu},
  pages = {36374--36411},
  title = {Internet of Agents: Weaving a Web of Heterogeneous Agents for Collaborative Intelligence},
  url = {https://proceedings.iclr.cc/paper_files/paper/2025/hash/59c27bf8d56d3d50c7aeaf7535dee975-Abstract-Conference.html},
  volume = {2025},
  year = {2025}
}

@inproceedings{chishti2026agentreputation,
  title = {AgentReputation: A Decentralized Agentic AI Reputation Framework},
  author = {Chishti, Mohd Sameen and Oyinloye, Damilare Peter and Li, Jingyue},
  booktitle = {Proceedings of the 34th ACM International Conference on the Foundations of Software Engineering},
  pages = {1317--1321},
  year = {2026},
  doi = {10.1145/3803437.3805579},
  url = {https://doi.org/10.1145/3803437.3805579}
}

@inproceedings{choi2025vote,
  title = {Debate or Vote: Which Yields Better Decisions in Multi-Agent Large Language Models?},
  author = {Choi, Hyeong Kyu and Zhu, Xiaojin and Li, Sharon},
  booktitle = {Advances in Neural Information Processing Systems},
  volume = {38},
  year = {2025},
  url = {https://proceedings.neurips.cc/paper_files/paper/2025/hash/934252acd87f254d5d4672fbde283bd2-Abstract-Conference.html},
  doi = {10.52202/085713-3405}
}

@misc{credence,
  title = {Credence: Real-time reputation scoring for {AI} agents},
  author = {{Credence Protocol}},
  howpublished = {\url{https://www.credenceprotocol.com/}},
  year = {2025},
  note = {Accessed: 2026-09-25.}
}

@inproceedings{cui2026freemad,
  title = {Free-mad: Consensus-free multi-agent debate},
  author = {Cui, Yu and Fu, Hang and Zhang, Haibin and Wang, Licheng and Zuo, Cong},
  booktitle = {Findings of the Association for Computational Linguistics: ACL 2026},
  pages = {31977--31997},
  year = {2026},
  doi = {10.18653/v1/2026.findings-acl.1600},
  url = {https://aclanthology.org/2026.findings-acl.1600/}
}

@inproceedings{dehghankar2025credibility,
    title = "An Adversary-Resistant Multi-Agent {LLM} System via Credibility Scoring",
    author = "Ebrahimi, Sana  and
      Dehghankar, Mohsen  and
      Asudeh, Abolfazl",
    editor = "Inui, Kentaro  and
      Sakti, Sakriani  and
      Wang, Haofen  and
      Wong, Derek F.  and
      Bhattacharyya, Pushpak  and
      Banerjee, Biplab  and
      Ekbal, Asif  and
      Chakraborty, Tanmoy  and
      Singh, Dhirendra Pratap",
    booktitle = "Proceedings of the 14th International Joint Conference on Natural Language Processing and the 4th Conference of the Asia-Pacific Chapter of the Association for Computational Linguistics",
    month = dec,
    year = "2025",
    address = "Mumbai, India",
    publisher = "The Asian Federation of Natural Language Processing and The Association for Computational Linguistics",
    url = "https://aclanthology.org/2025.ijcnlp-long.90/",
    doi = "10.18653/v1/2025.ijcnlp-long.90",
    pages = "1676--1693",
    ISBN = "979-8-89176-298-5"
}

@inproceedings{demszky2020goemotions,
  title={{GoEmotions}: A Dataset of Fine-Grained Emotions},
  author={Demszky, Dorottya and Movshovitz-Attias, Dana and Ko, Jeongwoo and Cowen, Alan and Nemade, Gaurav and Ravi, Sujith},
  booktitle={Proceedings of the 58th Annual Meeting of the Association for Computational Linguistics},
  year={2020}, pages={4040--4054}, publisher={Association for Computational Linguistics},
  doi={10.18653/v1/2020.acl-main.372},
  url={https://aclanthology.org/2020.acl-main.372/}
}

@article{deng2025agents,
author = {Deng, Zehang and Guo, Yongjian and Han, Changzhou and Ma, Wanlun and Xiong, Junwu and Wen, Sheng and Xiang, Yang},
title = {AI Agents Under Threat: A Survey of Key Security Challenges and Future Pathways},
year = {2025},
issue_date = {July 2025},
publisher = {Association for Computing Machinery},
address = {New York, NY, USA},
volume = {57},
number = {7},
issn = {0360-0300},
url = {https://doi.org/10.1145/3716628},
doi = {10.1145/3716628},
journal = {ACM Comput. Surv.},
month = feb,
articleno = {182},
numpages = {36}
}

@inproceedings{dettmers2023qlora,
  title = {{QLoRA}: Efficient Finetuning of Quantized {LLMs}},
  author = {Dettmers, Tim and Pagnoni, Artidoro and Holtzman, Ari and Zettlemoyer, Luke},
  booktitle = {Advances in Neural Information Processing Systems},
  volume = {36},
  year = {2023},
  url = {https://proceedings.neurips.cc/paper_files/paper/2023/hash/1feb87871436031bdc0f2beaa62a049b-Abstract-Conference.html}
}

@inproceedings{douceur2002sybil,
  author = {Douceur, John R.},
  editor = {Druschel, Peter
and Kaashoek, Frans
and Rowstron, Antony},
  title = {The Sybil Attack},
  booktitle = {Peer-to-Peer Systems},
  year = {2002},
  publisher = {Springer Berlin Heidelberg},
  address = {Berlin, Heidelberg},
  pages = {251--260},
  isbn = {978-3-540-45748-0},
  url = {https://www.microsoft.com/en-us/research/publication/the-sybil-attack/}
}

@inproceedings{du2024multiagent,
  title = {Improving Factuality and Reasoning in Language Models through Multiagent Debate},
  author = {Du, Yilun and Li, Shuang and Torralba, Antonio and Tenenbaum, Joshua B. and Mordatch, Igor},
  booktitle = {Proceedings of the 41st International Conference on Machine Learning},
  year = {2024},
  url = {https://proceedings.mlr.press/v235/du24e.html}
}

@misc{erc8004,
  title = {Ethereum Improvement Proposals. ERC-8004: Trustless Agents},
  author = {De Rossi, Marco and Crapis, Davide and Ellis, Jordan and Reppel, Erik},
  howpublished = {\url{https://eips.ethereum.org/EIPS/eip-8004}},
  year = {2025},
  note = {Draft specification. Accessed: 2026-09-25.}
}

@inproceedings{feldman2004robust,
author = {Feldman, Michal and Lai, Kevin and Stoica, Ion and Chuang, John},
title = {Robust incentive techniques for peer-to-peer networks},
year = {2004},
isbn = {1581137710},
publisher = {Association for Computing Machinery},
address = {New York, NY, USA},
url = {https://doi.org/10.1145/988772.988788},
doi = {10.1145/988772.988788},
booktitle = {Proceedings of the 5th ACM Conference on Electronic Commerce},
pages = {102–111},
numpages = {10},
location = {New York, NY, USA},
series = {EC '04}
}

@inproceedings{greshake2023not,
author = {Greshake, Kai and Abdelnabi, Sahar and Mishra, Shailesh and Endres, Christoph and Holz, Thorsten and Fritz, Mario},
title = {Not What You've Signed Up For: Compromising Real-World LLM-Integrated Applications with Indirect Prompt Injection},
year = {2023},
isbn = {9798400702600},
publisher = {Association for Computing Machinery},
address = {New York, NY, USA},
url = {https://doi.org/10.1145/3605764.3623985},
doi = {10.1145/3605764.3623985},
booktitle = {Proceedings of the 16th ACM Workshop on Artificial Intelligence and Security},
pages = {79–90},
numpages = {12},
location = {Copenhagen, Denmark},
series = {AISec '23}
}

@inproceedings{guo2017calibration,
  title = {On Calibration of Modern Neural Networks},
  author = {Guo, Chuan and Pleiss, Geoff and Sun, Yu and Weinberger, Kilian Q.},
  booktitle = {Proceedings of the 34th International Conference on Machine Learning},
  series = {Proceedings of Machine Learning Research},
  volume = {70},
  pages = {1321--1330},
  publisher = {PMLR},
  year = {2017},
  url = {https://proceedings.mlr.press/v70/guo17a.html}
}

@inproceedings{hendrycks2021math,
  title={Measuring Mathematical Problem Solving With the {MATH} Dataset},
  author={Hendrycks, Dan and Burns, Collin and Kadavath, Saurav and Arora, Akul and Basart, Steven and Tang, Eric and Song, Dawn and Steinhardt, Jacob},
  booktitle={Proceedings of the Neural Information Processing Systems Track on Datasets and Benchmarks},
  volume={1}, year={2021},
  url={https://datasets-benchmarks-proceedings.neurips.cc/paper/2021/hash/be83ab3ecd0db773eb2dc1b0a17836a1-Abstract-round2.html}
}

@inproceedings{herbrich2007trueskill,
  title={{TrueSkill}: A {Bayesian} Skill Rating System},
  author={Herbrich, Ralf and Minka, Tom and Graepel, Thore},
  booktitle={Advances in Neural Information Processing Systems},
  volume={19}, pages={569--576}, year={2006},
  url={https://proceedings.neurips.cc/paper/2006/hash/f44ee263952e65b3610b8ba51229d1f9-Abstract.html}
}

@article{hoffman2009survey,
  title = {A survey of attack and defense techniques for reputation systems},
  author = {Hoffman, Kevin and Zage, David and Nita-Rotaru, Cristina},
  journal = {ACM Computing Surveys (CSUR)},
  volume = {42},
  number = {1},
  pages = {1--31},
  year = {2009},
  publisher = {ACM New York, NY, USA},
  doi = {10.1145/1592451.1592452},
  url = {https://doi.org/10.1145/1592451.1592452}
}

@inproceedings{hu2022lora,
  title = {{LoRA}: Low-Rank Adaptation of Large Language Models},
  author = {Hu, Edward J. and Shen, Yelong and Wallis, Phillip and Allen-Zhu, Zeyuan and Li, Yuanzhi and Wang, Shean and Wang, Lu and Chen, Weizhu},
  booktitle = {International Conference on Learning Representations},
  year = {2022},
  url = {https://openreview.net/forum?id=nZeVKeeFYf9}
}

@inproceedings{hu2026lying,
  title = {Lying with truths: Open-channel multi-agent collusion for belief manipulation via generative montage},
  author = {Hu, Jinwei and Huang, Xinmiao and Sun, Youcheng and Dong, Yi and Huang, Xiaowei},
  booktitle = {Proceedings of the 64th Annual Meeting of the Association for Computational Linguistics (Volume 1: Long Papers)},
  pages = {5979--5996},
  year = {2026},
  doi = {10.18653/v1/2026.acl-long.270},
  url = {https://aclanthology.org/2026.acl-long.270/}
}

@inproceedings{hu2026multiagent,
  author = {Hu, Tianyu and Tan, Zhen and Wang, Song and Qu, Huaizhi and Chen, Tianlong},
  booktitle = {Advances in Neural Information Processing Systems},
  doi = {10.52202/085713-1548},
  editor = {D. Belgrave and C. Zhang and H. Lin and R. Pascanu and P. Koniusz and M. Ghassemi and N. Chen},
  pages = {46504--46540},
  publisher = {Curran Associates, Inc.},
  title = {Multi-Agent Debate for LLM Judges with Adaptive Stability Detection},
  url = {https://proceedings.neurips.cc/paper_files/paper/2025/hash/42475c537936b2394b5015e871765056-Abstract-Conference.html},
  volume = {38},
  year = {2025}
}

@article{huynh2006integrated,
  title = {An integrated trust and reputation model for open multi-agent systems},
  author = {Huynh, Trung Dong and Jennings, Nicholas R and Shadbolt, Nigel R},
  journal = {Autonomous agents and multi-agent systems},
  volume = {13},
  number = {2},
  pages = {119--154},
  year = {2006},
  publisher = {Springer},
  doi = {10.1007/s10458-005-6825-4},
  url = {https://link.springer.com/article/10.1007/s10458-005-6825-4}
}

@article{irving2018ai,
  title = {AI safety via debate},
  author = {Irving, Geoffrey and Christiano, Paul and Amodei, Dario},
  journal = {arXiv preprint arXiv:1805.00899},
  year = {2018},
  url = {https://arxiv.org/abs/1805.00899}
}

@inproceedings{josang2002beta,
  title={The Beta Reputation System},
  author={J{\o}sang, Audun and Ismail, Roslan},
  booktitle={Proceedings of the 15th Bled Electronic Commerce Conference},
  year={2002},
  url={https://aisel.aisnet.org/bled2002/41/}
}

@inproceedings{just2014mutants,
  title = {Are mutants a valid substitute for real faults in software testing?},
  author = {Just, Ren{\'e} and Jalali, Darioush and Inozemtseva, Laura and Ernst, Michael D and Holmes, Reid and Fraser, Gordon},
  booktitle = {Proceedings of the 22nd ACM SIGSOFT international symposium on foundations of software engineering},
  pages = {654--665},
  year = {2014},
  doi = {10.1145/2635868.2635929},
  url = {https://doi.org/10.1145/2635868.2635929}
}

@inproceedings{kaesberg2025voting,
  title = {Voting or consensus? decision-making in multi-agent debate},
  author = {Kaesberg, Lars Benedikt and Becker, Jonas and Wahle, Jan Philip and Ruas, Terry and Gipp, Bela},
  booktitle = {Findings of the Association for Computational Linguistics: ACL 2025},
  pages = {11640--11671},
  year = {2025},
  doi = {10.18653/v1/2025.findings-acl.606},
  url = {https://aclanthology.org/2025.findings-acl.606/}
}

@inproceedings{kamvar2003eigentrust,
  title={The {EigenTrust} Algorithm for Reputation Management in {P2P} Networks},
  author={Kamvar, Sepandar D. and Schlosser, Mario T. and Garcia-Molina, Hector},
  booktitle={Proceedings of the 12th International Conference on World Wide Web},
  pages={640--651}, year={2003}, publisher={ACM},
  doi={10.1145/775152.775242},
  url={https://dl.acm.org/doi/10.1145/775152.775242}
}

@inproceedings{karandikar2021soft,
  title = {Soft Calibration Objectives for Neural Networks},
  author = {Karandikar, Archit and Cain, Nicholas and Tran, Dustin and Lakshminarayanan, Balaji and Shlens, Jonathon and Mozer, Michael and Roelofs, Becca},
  booktitle = {Advances in Neural Information Processing Systems},
  volume = {34},
  year = {2021},
  url = {https://proceedings.neurips.cc/paper_files/paper/2021/hash/f8905bd3df64ace64a68e154ba72f24c-Abstract.html}
}

@inproceedings{kavathekar-etal-2026-tamas,
  title = {{TAMAS}: Benchmarking Adversarial Risks in Multi-Agent {LLM} Systems},
  author = {Kavathekar, Ishan and Jain, Hemang and Rathod, Ameya and
          Kumaraguru, Ponnurangam and Ganu, Tanuja},
  booktitle = {Proceedings of the 64th Annual Meeting of the Association for Computational Linguistics (Volume 1: Long Papers)},
  year = {2026},
  pages = {31238--31268},
  url = {https://aclanthology.org/2026.acl-long.1442/},
  doi = {10.18653/v1/2026.acl-long.1442}
}

@inproceedings{khan2024debating,
  title={Debating with More Persuasive {LLM}s Leads to More Truthful Answers},
  author={Khan, Akbir and Hughes, John and Valentine, Dan and Ruis, Laura and Sachan, Kshitij and Radhakrishnan, Ansh and Grefenstette, Edward and Bowman, Samuel R. and Rockt{\"a}schel, Tim and Perez, Ethan},
  booktitle={Proceedings of the 41st International Conference on Machine Learning},
  series={Proceedings of Machine Learning Research},
  volume={235}, pages={23662--23733}, year={2024}, publisher={PMLR},
  url={https://proceedings.mlr.press/v235/khan24a.html}
}

@inproceedings{klos2003decentralized,
  title = {Decentralized reputation-based trust for assessing agent reliability under aggregate feedback},
  author = {Klos, Tomas B and La Poutr{\'e}, Han},
  booktitle = {Trusting Agents for Trusting Electronic Societies},
  pages = {110--128},
  year = {2005},
  organization = {Springer},
  series = {Lecture Notes in Computer Science},
  volume = {3577},
  doi = {10.1007/11532095_7},
  url = {https://link.springer.com/chapter/10.1007/11532095_7}
}

@inproceedings{ko2026social,
  title = {Social Dynamics as Critical Vulnerabilities that Undermine Objective Decision-Making in LLM Collectives},
  author = {Ko, Changgeon and Shin, Jisu and Song, Hoyun and Lee, Huije and Hwang, Eui Jun and Park, Jong C},
  booktitle = {Proceedings of the 64th Annual Meeting of the Association for Computational Linguistics (Volume 1: Long Papers)},
  pages = {37865--37890},
  year = {2026},
  doi = {10.18653/v1/2026.acl-long.1756},
  url = {https://aclanthology.org/2026.acl-long.1756/}
}

@article{koutrouli2012taxonomy,
  title = {Taxonomy of attacks and defense mechanisms in P2P reputation systems—Lessons for reputation system designers},
  journal = {Computer Science Review},
  volume = {6},
  number = {2--3},
  pages = {47--70},
  year = {2012},
  issn = {1574-0137},
  doi = {10.1016/j.cosrev.2012.01.002},
  url = {https://www.sciencedirect.com/science/article/pii/S1574013712000093},
  author = {Eleni Koutrouli and Aphrodite Tsalgatidou}
}

@article{kraidia2026collaboration,
  title = {When collaboration fails: persuasion driven adversarial influence in multi agent large language model debate},
  author = {Kraidia, Insaf and Qaddara, Iyas and Almutairi, Alhanof and Alzaben, Nada and Belhouari, Samir Brahim},
  journal = {Scientific Reports},
  volume = {16},
  number = {1},
  pages = {11640},
  year = {2026},
  publisher = {Nature Publishing Group UK London},
  doi = {10.1038/s41598-026-42705-7},
  url = {https://www.nature.com/articles/s41598-026-42705-7}
}

@misc{lee2026trueskillsoftware,
  title = {{TrueSkill}: Python Implementation Documentation},
  author = {Lee, Heungsub},
  howpublished = {Software documentation, version 0.4.5},
  note = {Accessed September 25, 2026},
  year = {2026},
  url = {https://trueskill.org/}
}

@article{li2024more,
  title = {More Agents Is All You Need},
  author = {Junyou Li and Qin Zhang and Yangbin Yu and Qiang Fu and Deheng Ye},
  journal = {Transactions on Machine Learning Research},
  issn = {2835-8856},
  year = {2024},
  url = {https://openreview.net/forum?id=bgzUSZ8aeg},
  note = {}
}

@inproceedings{lin2017focal,
  title = {Focal Loss for Dense Object Detection},
  author = {Lin, Tsung-Yi and Goyal, Priya and Girshick, Ross and He, Kaiming and Doll{\'a}r, Piotr},
  booktitle = {2017 IEEE International Conference on Computer Vision (ICCV)},
  pages = {2999--3007},
  year = {2017},
  doi = {10.1109/ICCV.2017.324},
  url = {https://doi.org/10.1109/ICCV.2017.324}
}

@article{lin2026too_many_agents,
  author  = {Lin, Belle},
  title   = {Companies Have a New AI Problem: Too Many Agents},
  journal = {The Wall Street Journal},
  year    = {2026},
  month   = may,
  day     = {15},
  url     = {https://www.wsj.com/cio-journal/companies-have-a-new-ai-problem-too-many-agents-9539c4d6},
  note    = {Accessed: September 17, 2026}
}

@inproceedings{liu2025can,
  title = {Can an Individual Manipulate the Collective Decisions of Multi-Agents?},
  author = {Liu, Fengyuan and Zhao, Rui and Chen, Shuo and Li, Guohao and Torr, Philip and Han, Lei and Gu, Jindong},
  booktitle = {Proceedings of the 2025 Conference on Empirical Methods in Natural Language Processing},
  pages = {12158--12182},
  year = {2025},
  doi = {10.18653/v1/2025.emnlp-main.611},
  url = {https://aclanthology.org/2025.emnlp-main.611/}
}

@inproceedings{lou2025drf,
  title = {{DRF}: {LLM-AGENT} Dynamic Reputation Filtering Framework},
  author = {Lou, Yuwei and Hu, Hao and Ma, Shaocong and Zhang, Zongfei and Wang, Liang and Ge, Jidong and Tao, Xianping},
  booktitle = {Neural Information Processing},
  pages = {127--141},
  year = {2026},
  organization = {Springer},
  series = {Lecture Notes in Computer Science},
  volume = {16312},
  doi = {10.1007/978-981-95-4384-7_10},
  url = {https://link.springer.com/chapter/10.1007/978-981-95-4384-7_10}
}

@inproceedings{mui2002notions,
  title = {Notions of reputation in multi-agents systems: a review},
  author = {Mui, Lik and Mohtashemi, Mojdeh and Halberstadt, Ari},
  booktitle = {Proceedings of the first international joint conference on Autonomous agents and multiagent systems: part 1},
  pages = {280--287},
  year = {2002},
  doi = {10.1145/544741.544807},
  url = {https://doi.org/10.1145/544741.544807}
}

@article{park2026data,
  author = {Park, Jeongsu and Lim, Yuji and Kim, Geonwoo and Yun, Taehyeon and Min, Moohong},
  title = {Data poisoning in LLM multiagent societies: Social proof drives collective decision failure in financial deliberation},
  journal = {ETRI Journal},
  year = {2026},
  doi = {10.4218/etrij.2026-0186},
  pages = {1--13},
  url = {https://onlinelibrary.wiley.com/doi/10.4218/etrij.2026-0186}
}

@inproceedings{petrovic2018state,
  title = {State of mutation testing at google},
  author = {Petrovi{\'c}, Goran and Ivankovi{\'c}, Marko},
  booktitle = {Proceedings of the 40th international conference on software engineering: Software engineering in practice},
  pages = {163--171},
  year = {2018},
  url = {https://research.google/pubs/state-of-mutation-testing-at-google/}
}

@article{qwen2025qwen25,
  title = {{Qwen2.5} Technical Report},
  author = {{Qwen} and Yang, An and Yang, Baosong and Zhang, Beichen and Hui, Binyuan and Zheng, Bo and Yu, Bowen and Li, Chengyuan and Liu, Dayiheng and Huang, Fei and Wei, Haoran and Lin, Huan and Yang, Jian and Tu, Jianhong and Zhang, Jianwei and Yang, Jianxin and Yang, Jiaxi and Zhou, Jingren and Lin, Junyang and Dang, Kai and Lu, Keming and Bao, Keqin and Yang, Kexin and Yu, Le and Li, Mei and Xue, Mingfeng and Zhang, Pei and Zhu, Qin and Men, Rui and Lin, Runji and Li, Tianhao and Tang, Tianyi and Xia, Tingyu and Ren, Xingzhang and Ren, Xuancheng and Fan, Yang and Su, Yang and Zhang, Yichang and Wan, Yu and Liu, Yuqiong and Cui, Zeyu and Zhang, Zhenru and Qiu, Zihan},
  journal = {arXiv preprint arXiv:2412.15115},
  year = {2025},
  note = {Version 2},
  doi = {10.48550/arXiv.2412.15115},
  url = {https://arxiv.org/abs/2412.15115}
}

@inproceedings{ren2025reputation,
  title = {Reputation as a Solution to Cooperation Collapse in {LLM}-based {MASs}},
  author = {Ren, Siyue and Fu, Wanli and Zou, Xinkun and Shen, Chen and Cai, Yi and Chu, Chen and Wang, Zhen and Hu, Shuyue},
  booktitle = {Proceedings of the 25th International Conference on Autonomous Agents and Multiagent Systems},
  year = {2026},
  pages = {245--253},
  publisher = {International Foundation for Autonomous Agents and Multiagent Systems},
  url = {https://doi.org/10.65109/UEHN4980},
  doi = {10.65109/UEHN4980}
}

@incollection{resnick2002trust,
  author = {Resnick, Paul and Zeckhauser, Richard},
  editor = {Baye, Michael R.},
  isbn = {978-0-76230-971-9},
  title = {Trust among strangers in internet transactions: Empirical analysis of {eBay}'s reputation system},
  booktitle = {The Economics of the Internet and E-commerce},
  publisher = {Emerald Group Publishing Limited},
  year = {2002},
  month = {10},
  doi = {10.1016/S0278-0984(02)11030-3},
  url = {https://doi.org/10.1016/S0278-0984(02)11030-3},
  eprint = {https://www.emerald.com/book/chapter-pdf/9084920/s0278-0984_02_11030-3.pdf},
  pages = {127--157}
}

@article{rumelhart1986learning,
  title = {Learning representations by back-propagating errors},
  author = {Rumelhart, David E and Hinton, Geoffrey E and Williams, Ronald J},
  journal = {Nature},
  volume = {323},
  number = {6088},
  pages = {533--536},
  year = {1986},
  publisher = {Nature Publishing Group UK London},
  doi = {10.1038/323533a0},
  url = {https://www.nature.com/articles/323533a0}
}

@misc{said,
  title = {{SAID}: The identity and reputation layer for {AI} agents},
  author = {{SAID Protocol}},
  howpublished = {\url{https://www.saidprotocol.com/}},
  year = {2025},
  note = {Accessed: 2026-09-25.}
}

@article{shapira2026agents,
  title = {Agents of chaos},
  author = {Shapira, Natalie and Wendler, Chris and Yen, Avery and Sarti, Gabriele and Pal, Koyena and Floody, Olivia and Belfki, Adam and Loftus, Alex and Jannali, Aditya Ratan and Prakash, Nikhil and others},
  journal = {arXiv preprint arXiv:2602.20021},
  year = {2026},
  url = {https://arxiv.org/abs/2602.20021}
}

@article{sokolova2009systematic,
  title = {A systematic analysis of performance measures for classification tasks},
  author = {Sokolova, Marina and Lapalme, Guy},
  journal = {Information Processing \& Management},
  volume = {45},
  number = {4},
  pages = {427--437},
  year = {2009},
  publisher = {Elsevier},
  doi = {10.1016/j.ipm.2009.03.002},
  url = {https://www.sciencedirect.com/science/article/pii/S0306457309000259}
}

@inproceedings{srivatsa2005trustguard,
author = {Srivatsa, Mudhakar and Xiong, Li and Liu, Ling},
title = {TrustGuard: countering vulnerabilities in reputation management for decentralized overlay networks},
year = {2005},
isbn = {1595930469},
publisher = {Association for Computing Machinery},
address = {New York, NY, USA},
url = {https://doi.org/10.1145/1060745.1060808},
doi = {10.1145/1060745.1060808},
booktitle = {Proceedings of the 14th International Conference on World Wide Web},
pages = {422–431},
numpages = {10},
location = {Chiba, Japan},
series = {WWW '05}
}

@inproceedings{
wang2023selfconsistency,
title={Self-Consistency Improves Chain of Thought Reasoning in Language Models},
author={Xuezhi Wang and Jason Wei and Dale Schuurmans and Quoc V Le and Ed H. Chi and Sharan Narang and Aakanksha Chowdhery and Denny Zhou},
booktitle={The Eleventh International Conference on Learning Representations },
year={2023},
url={https://openreview.net/forum?id=1PL1NIMMrw}
}

@article{wang2023survey,
  title = {A survey on large language model based autonomous agents},
  author = {Wang, Lei and Ma, Chen and Feng, Xueyang and Zhang, Zeyu and Yang, Hao and Zhang, Jingsen and Chen, Zhiyuan and Tang, Jiakai and Chen, Xu and Lin, Yankai and others},
  journal = {Frontiers of computer science},
  volume = {18},
  number = {6},
  pages = {186345},
  year = {2024},
  publisher = {Springer},
  doi = {10.1007/s11704-024-40231-1},
  url = {https://link.springer.com/article/10.1007/s11704-024-40231-1}
}

@article{xiong2026trustless,
  title = {Can Trustless Agents Be Trusted? An Empirical Study of the {ERC-8004} Decentralized {AI} Agent Ecosystem},
  author = {Xiong, Xihan and Li, Zelin and Wei, Wei and Wang, Qin and Knottenbelt, William and Wang, Zhipeng},
  journal = {arXiv preprint arXiv:2606.26028},
  year = {2026},
  url = {https://arxiv.org/abs/2606.26028}
}

@inproceedings{yang2024watch,
  author = {Yang, Wenkai and Bi, Xiaohan and Lin, Yankai and Chen, Sishuo and Zhou, Jie and Sun, Xu},
  booktitle = {Advances in Neural Information Processing Systems},
  doi = {10.52202/079017-3201},
  editor = {A. Globerson and L. Mackey and D. Belgrave and A. Fan and U. Paquet and J. Tomczak and C. Zhang},
  pages = {100938--100964},
  publisher = {Curran Associates, Inc.},
  title = {Watch Out for Your Agents! Investigating Backdoor Threats to LLM-Based Agents},
  url = {https://proceedings.neurips.cc/paper_files/paper/2024/hash/b6e9d6f4f3428cd5f3f9e9bbae2cab10-Abstract-Conference.html},
  volume = {37},
  year = {2024}
}

@inproceedings{yu2025humanevalpro,
  title={{HumanEval Pro} and {MBPP Pro}: Evaluating Large Language Models on Self-invoking Code Generation Task},
  author={Yu, Zhaojian and Zhao, Yilun and Cohan, Arman and Zhang, Xiao-Ping},
  booktitle={Findings of the Association for Computational Linguistics: ACL 2025},
  year={2025}, pages={13253--13279}, publisher={Association for Computational Linguistics},
  doi={10.18653/v1/2025.findings-acl.686},
  url={https://aclanthology.org/2025.findings-acl.686/}
}

@inproceedings{zhang2025agent,
  title = {Agent security bench (asb): Formalizing and benchmarking attacks and defenses in llm-based agents},
  author = {Zhang, Hanrong and Huang, Jingyuan and Mei, Kai and Yao, Yifei and Wang, Zhenting and Zhan, Chenlu and Wang, Hongwei and Zhang, Yongfeng},
  booktitle = {International Conference on Learning Representations},
  volume = {2025},
  pages = {35331--35366},
  year = {2025},
  url = {https://openreview.net/forum?id=V4y0CpX4hK}
}

@inproceedings{zheng2026rethinking,
  title = {Rethinking the Reliability of Multi-agent System: A Perspective from {Byzantine} Fault Tolerance},
  author = {Zheng, Lifan and Chen, Jiawei and Yin, Qinghong and Zhang, Jingyuan and Zeng, Xinyi and Tian, Yu},
  booktitle = {Proceedings of the AAAI Conference on Artificial Intelligence},
  volume = {40},
  pages = {35012--35020},
  year = {2026},
  doi = {10.1609/aaai.v40i41.40806},
  url = {https://ojs.aaai.org/index.php/AAAI/article/view/40806}
}

\clearpage
\appendix
\section{Notations and Baselines}
\label{app:notation}

\subsection{Notations}
\tabref{tab:notation} collects the symbols used in \secrref{sec:model}, \secrref{sec:design}, and the detailed definitions below. Identities are indexed by $i$ and tasks by $t$. Per-task quantities carry a task subscript only where the task matters.
Task $t$ reads $\state_{t-1}$ and writes $\state_t$ after the answer;
method-specific quantities use local superscripts where needed.

\begin{table}[!htbp]
\centering
\caption{Core notation.}
\label{tab:notation}
\footnotesize
\setlength{\tabcolsep}{5pt}
\renewcommand{\arraystretch}{1.05}
\begin{tabularx}{\linewidth}{lX}
\toprule
Symbol & Meaning \\
\midrule
\multicolumn{2}{@{}l}{\textit{Protocol related}} \\
$\agents$; $\nagents$; $\nfaulty$ & Identity set; number of identities; maximum adversarial identities. \\
$\task = (\ctx, \param)$ & Task: context $\ctx$ (statement and public evidence) and parameter $\param$. \\
$\proposal_i = (\cont_i, \reason_i)$ & Response of identity $i$: payload $\cont_i$ and rationale $\reason_i$. \\
$\ndebate$ & Debate rounds ($\ndebate = 0$ in the main setting). \\
$G$; clone-group mapping & Known group of identities; the experiments group replicas by their underlying API model. \\
$\adv$; $\coalition_t$ & Adversary operator; coalition active on task $t$, $|\coalition_t| \le \nfaulty$. \\
$P_t$; $k$ & Identities selected on a task; target pool size. \\
$\aggfunc$; $\state$ & Aggregation rule (the \decide step); persistent reputation state. \\
\midrule
\multicolumn{2}{@{}l}{\textit{Same-task signals}} \\
$p^{pub}_{i,t}$ & Public support (Eq.~\ref{eq:public-support}); distinct from the task context $\ctx$. \\
$p^{sem}_{i,t}$ & Calibrated correctness probability from the 1.5B semantic verifier. \\
$r^{atk}_{i,t}$ & Behavior-model risk score (Appendix~\ref{app:design:feedback}). \\
$\eta_{i,t}$; $f_{i,t}$ & Parse-validity indicator; shared detector flag. The response remains $\proposal_{i,t}$. \\
$a_{i,t}$; $r^{gate}_{i,t}$ & Fused evidence score; derived decision penalty, not a second learned probability (Eq.~\ref{eq:gate-risk}). \\
$\widehat a_{i,t}$ & Alarm strength used in persistent updates: $\max\{a_{i,t},0.90d_{i,t}\}$. \\
$q_t^{med}$; $s^{self}_{i,t}$ & Median parse-valid quality; same-model self-consistency. \\
$c^{fmt}_{i,t}$ & Format compliance in $[0,1]$; parse failure forces $q_{i,t}=0$. \\
$s_{i,t}$; $q_{i,t}$ & Fused support; same-task quality (Appendix~\ref{app:design:fusion}). \\
$d_{i,t}$ & Current-task block: parse failure or the shared detector flag $f_{i,t}$ (Eq.~\ref{eq:behavior-alarm}). \\
blocked set & Current-task blocked identities or identities with active temporary restrictions. They rank behind all unblocked candidates. \\
\midrule
\multicolumn{2}{@{}l}{\textit{Weights and selection}} \\
$w^{hist}_{i,t}$ & Incoming weight from $\state_{t-1}$, with numerical floor (Eq.~\ref{eq:hist-weight}). \\
$w_{i,t}$; $Q_{i,t}$ & Final pool weight; historical-risk/cooldown mark used in current weighting. \\
$\widetilde w_{i,t}$; $m_{i,t}$ & Pre-group current-round weight (Eq.~\ref{eq:current-weight}); product of blocking, historical-risk and invalidity multipliers. \\
$M_G$; $p_G$ & Pre-group weight sum; largest member weight in $G$, with $m=|G|$. \\
\midrule
\multicolumn{2}{@{}l}{\textit{Persistent state}} \\
$\bar q_i$; $q^{short}_i$ & Long-term quality (Beta-style prior, strength 12); short-term quality (asymmetric EMA $0.30/0.08$). \\
$\bar e_i$; $e^{short}_i$ & Long- and short-term correctness support. \\
$D_i$; $K_i$; $E_i$; $I_i$ & Failure debt; strike counter; detector EMA; identity risk (Appendix~\ref{app:design:identity}). \\
$r^{temp}_i$ & Temporal risk: maximum of $D_i$, level risk, $E_i$, $I_i$, and instant failure risk. \\
$h_i$ & Cooldown counter (three tasks after a valid-response detection). \\
\midrule
\multicolumn{2}{@{}l}{\textit{Baseline and evaluation quantities}} \\
$y^{cand}_{i,t}$; $F_t$; $\mathcal F_t$ & Parsed candidate answer; feedback-available identities; detector-selected candidates. \\
$z_{i,t}$; $u_{i,t}$; $x_{i,t}$ & Estimated feedback quality; validity-masked score; format-adjusted score. \\
$\sigma^{\rm SM}_{i,t}$; $h_i^{\rm TS}$ & Single-Metric quality EMA; TrueSkill conservative rating. \\
$S^{\mathrm{Bab}}_{i,t}$; $B_{i,t}$ & Babylon score; earlier-performance baseline for gap tracking (Appendix~\ref{app:design:histweight}). \\
$A_t$; $J_t$; $E_t$; $H_t$ & Active-attack indicator; payload-bearing identities; exposure and payload-match indicators. \\
$s^0_{m,t}$; $s^a_{m,t}$ & Clean and attacked per-task utility for method $m$. \\
\bottomrule
\end{tabularx}
\end{table}

\subsection{Baselines} \label{app:baseline}

We distinguish source-method defaults from settings introduced by our MAD adaptations; feedback mappings, clipping bounds, and numerical tolerances are implementation choices unless stated otherwise.

\heading{Multi-agent debate, formally.}
We formalize the MAD phases introduced in \secrref{sec:model}, using the notation in \tabref{tab:notation}. Consider a sequence
of tasks $\task_t=(\ctx_t,\param_t)$ indexed by $t=1,2,\ldots$. In the
\propose phase, each identity $i\in\agents$ independently generates
\begin{equation}
\proposal_{i,t}^{(0)}
=
\operatorname{Propose}_i(\task_t).
\label{eq:jm-propose}
\end{equation}
If the protocol includes debate, then at each round
$\ell=1,\ldots,\ndebate$, identity $i$ updates its proposal according to
\begin{equation}
\proposal_{i,t}^{(\ell)}
=
\operatorname{Debate}_i
\left(
\task_t,
\proposal_{i,t}^{(\ell-1)},
\mathcal{H}_{i,t}^{(\ell-1)}
\right),
\label{eq:jm-debate}
\end{equation}
where $\mathcal{H}_{i,t}^{(\ell-1)}$ contains the proposals and messages
visible to identity $i$ before round $\ell$. The MAD communication topology
determines the contents of this history. We use
\begin{equation}
\proposal_{i,t}
=
\proposal_{i,t}^{(\ndebate)}
=
(\cont_{i,t},\reason_{i,t})
\end{equation}
to denote the final proposal submitted to the \decide phase. When
$\ndebate=0$, as in our main setting, this is simply the initial proposal
$\proposal_{i,t}^{(0)}$.

The \decide phase applies an aggregation rule $\aggfunc$ to a selected pool
$P_t\subseteq\agents$. For the voting-based methods considered below, let
$y^{cand}_{i,t}$ denote the normalized answer parsed from the payload
$\cont_{i,t}$. We set $y^{cand}_{i,t}=\bot$ if parsing fails. Given nonnegative
weights $w_{i,t}$ for the selected identities, the final answer is
\begin{equation}
\hat y_t
\in
\arg\max_{a\neq\bot}
\sum_{i\in P_t}
w_{i,t}\mathbb{I}[y^{cand}_{i,t}=a].
\label{eq:jm-common-aggregation}
\end{equation}
\equaref{eq:jm-common-aggregation} gives weighted answer-group
voting for MATH, with equality interpreted through the protocol's candidate
comparison. Label-set and code readouts are specified under task-dependent
aggregation below.

To compare aggregation methods independently of LLM-generation variance, we
generate and cache the final proposal set
\begin{equation}
\mathcal{R}_t
=
\{\proposal_{i,t}:i\in\agents\}
\end{equation}
once for each task. Every aggregation method receives the same
$\mathcal{R}_t$. Thus, any difference in the resulting decision is caused by
which agents are included in $P_t$, how much weight is assigned to each selected proposal, and how the weighted proposals are combined into the final answer,
rather than by different sampled agent responses.

\heading{Estimated-quality feedback.}
To support the stateful methods evaluated in \secrref{sec:evaluation:setup}, we define an estimated-quality feedback interface. This interface is introduced for our
experimental comparison and is not a standard component of MAD protocols.
For each identity $i$ whose proposal is observed on task $t$, let
$z_{i,t}\in[0,1]$ denote an estimated quality score, $\eta_{i,t}\in\{0,1\}$
indicate whether the proposal can be parsed successfully, and
$c^{fmt}_{i,t}\in[0,1]$ measure format compliance. We define
\[
u_{i,t}=\eta_{i,t}z_{i,t},
\qquad
x_{i,t}=z_{i,t}(0.75+0.25c^{fmt}_{i,t}).
\]
These quantities provide estimated-quality signals rather than benchmark
ground-truth labels. Stateful aggregation methods update their persistent
state only after the decision on task $t$ has been made.

\heading{Uniform Majority and Uniform Random-$k$.}
Neither Uniform Majority nor Uniform Random-$k$ maintains persistent state.
Uniform Majority applies equal-weight voting to all identities
~\citep{kaesberg2025voting}:
\begin{equation}
P_t=\agents,
\qquad
w_{i,t}=1
\quad(i\in\agents).
\label{eq:jm-uniform-majority}
\end{equation}
Its pool therefore contains all $\nagents=10$ identities. This baseline shows the result when every agent participates and has the same influence, regardless of its performance on previous tasks.

Uniform Random-$k$ is our size-matched sampling control. It samples a
size-$k$ pool uniformly without replacement and assigns equal weight to
every selected identity:
\begin{equation}
\Pr(P_t=S)
=
\binom{\nagents}{k}^{-1}
\quad
\text{for every }S\subseteq\agents\text{ with }|S|=k,
\qquad
w_{i,t}=1
\quad(i\in P_t).
\label{eq:jm-uniform-random-k}
\end{equation}
The selection is independent of $\state$, agent feedback, and past
performance. To make the control reproducible, the implementation uses a
fixed pseudorandom seed for each experimental run and combines it with the
task identifier when sampling $P_t$. Repeating the same run therefore yields
the same pool for every task. This baseline isolates the effect of using only
$k$ identities without using reputation for selection or weighting.

For the stateful top-$k$ methods considered below, $P_t$ instead contains
the $k$ identities with the largest weights available before task $t$.
Weight ties are resolved using a deterministic seeded rule. During the
configured all-identity warm-up period, these baselines use
$P_t=\agents$. In contrast, \rep applies its same-task screening mechanism
and constructs a size-$k$ pool even during warm-up.

\heading{Single-Metric.}
Single-Metric is an in-study control designed to test what can be achieved
using only one historical quality indicator per identity. It excludes the
additional mechanisms used by \rep, including identity-risk detection,
clone-group caps, cooldown, and same-task screening.

Let $\sigma^{\mathrm{SM}}_{i,t}$ denote the exponentially smoothed historical
quality of identity $i$ after incorporating feedback available through task
$t$. This symbol is distinct from the fused same-task support $s_{i,t}$
defined in \tabref{tab:notation}. We initialize
\begin{equation}
\sigma^{\mathrm{SM}}_{i,0}=0.5
\quad(i\in\agents).
\end{equation}
Before task $t$, Single-Metric uses the state available through task $t-1$
to assign
\begin{equation}
w_{i,t}
=
\operatorname{clip}_{[0.05,\,1.50]}
\left(\sigma^{\mathrm{SM}}_{i,t-1}\right).
\label{eq:jm-single-metric-weight}
\end{equation}
It selects the $k$ identities with the largest $w_{i,t}$ values and applies
the weighted-vote rule in \equaref{eq:jm-common-aggregation}. The same scalar
historical score therefore controls both agent selection and voting weight.

After feedback for task $t$ becomes available, the score of every observed
identity is updated using a fixed-rate exponential moving average.
Let $F_t$ denote the identities for which feedback is available for this
update (distinct from the detector's candidate set $\mathcal F_t$):
\begin{equation}
\sigma^{\mathrm{SM}}_{i,t}
=
0.85\sigma^{\mathrm{SM}}_{i,t-1}
+
0.15x_{i,t}
\quad(i\in F_t).
\label{eq:jm-single-metric-update}
\end{equation}
For an unobserved identity, the previous score is carried forward:
\begin{equation}
\sigma^{\mathrm{SM}}_{i,t}
=
\sigma^{\mathrm{SM}}_{i,t-1}
\quad(i\notin F_t).
\label{eq:jm-single-metric-unobserved}
\end{equation}
The update coefficient $0.15$ controls how quickly recent feedback changes
the historical estimate. Single-Metric otherwise maintains no additional
persistent signals or current-task controls. Its initialization, clipping interval, and update rate are fixed settings of this in-study control.

The four reputation baselines below receive the same quality feedback $u_{i,t}$ after task $t$. Each baseline uses this feedback to update the weight $w_{i,t+1}$ for the next task. Unlike \rep, these baselines do not use the current proposal to adjust an agent's weight before the current answer is produced.

\heading{Babylon.} We adapt Babylon~\citep{babylon2026software} to maintain one reputation score for each agent. Let $n_i$ be the number of observed tasks, and let
\begin{equation}
W_i
=
\frac{1}{n_i}
\sum_{\tau}
\mathbf{1}[z_{i,\tau}\ge0.999]
\end{equation}
be the fraction of observations with a nearly perfect verifier score. The adaptation also maintains $F_i$ and $J_i$ as running averages of
\begin{equation}
(20+80z_{i,t})
\left(
0.75+0.25c^{fmt}_{i,t}
\right).
\end{equation}
Thus, both records increase with the verifier score and with compliance with the required output format. A separate record $V_{i,t}$ tracks the verifier score over time:
\begin{equation}
V_{i,0}=0.5,
\qquad
V_{i,t}
=
0.90V_{i,t-1}
+
\frac{0.10}{1+\exp[-(2z_{i,t}-1)]}.
\end{equation}
Babylon combines these records with the number of observed tasks:
\begin{equation}
\begin{aligned}
S^{\mathrm{Bab}}_{i,t}
={}&
0.4
\left[
100
\left(
0.7V_{i,t}+0.3W_i
\right)
\right]
+
0.4
\left(
0.7F_i+0.3J_i
\right)\\
&+
0.2\min\{100,2n_i\},\\
w_{i,t+1}
={}&
\max\{0.05,S^{\mathrm{Bab}}_{i,t}/100\}.
\end{aligned}
\label{eq:jm-babylon}
\end{equation}
The initial reputation is $50$. The $0.4/0.4/0.2$ component weights, $0.7/0.3$ blends, activity term, and initial score follow the cited implementation; the verifier-based feedback mapping is our MAD adaptation. This baseline evaluates whether a single score combining past performance, format compliance, and the number of observed tasks is sufficient for MAD aggregation.

\heading{Beta Reputation.} Beta Reputation~\citep{josang2002beta} accumulates positive and negative evidence for each agent. The quality feedback $u_{i,t}$ is treated as fractional positive evidence, and $1-u_{i,t}$ is treated as fractional negative evidence:
\begin{equation}
\alpha_{i,0}
=
\beta_{i,0}
=
1,
\qquad
\alpha_{i,t}
=
\alpha_{i,t-1}
+
u_{i,t},
\qquad
\beta_{i,t}
=
\beta_{i,t-1}
+
1-u_{i,t}.
\end{equation}
The weight for the next task is proportional to the estimated fraction of positive evidence:
\begin{equation}
w_{i,t+1}
\propto
\frac{\alpha_{i,t}}
{\alpha_{i,t}+\beta_{i,t}}.
\label{eq:jm-beta-baseline}
\end{equation}
The weights are normalized across agents. All previous observations remain in $\alpha_i$ and $\beta_i$, so older evidence is not discarded or reduced. The $(1,1)$ initialization corresponds to a uniform Beta prior. This baseline evaluates direct accumulation of quality evidence without maintaining a separate record of suspicious behavior.

\heading{EigenTrust.} EigenTrust~\citep{kamvar2003eigentrust} originally computes trust from ratings between participants. Our setting does not collect pairwise ratings from agents. We therefore use the shared quality feedback $u_{j,t}$ as evidence about each observed agent $j$. For every pair $(i,j)$, we update
\begin{equation}
S^+_{ij}
\leftarrow
S^+_{ij}
+
u_{j,t},
\qquad
S^-_{ij}
\leftarrow
S^-_{ij}
+
1-u_{j,t},
\end{equation}
with $S^+_{ij}=S^-_{ij}=0$ initially. Because the same feedback is available to the system, every row receives the same evidence about agent $j$. We define
\begin{equation}
s_{ij}
=
\max\{S^+_{ij}-S^-_{ij},0\},
\end{equation}
and normalize each row:
\begin{equation}
C_{ij}
=
\begin{cases}
s_{ij}/\sum_\ell s_{i\ell},
&
\sum_\ell s_{i\ell}>0,\\
1/\nagents,
&
\text{otherwise}.
\end{cases}
\end{equation}
The weights are obtained by repeatedly applying
\begin{equation}
\boldsymbol w
=
0.85C^{\mathsf T}\boldsymbol w
+
0.15\boldsymbol p,
\qquad
p_i
=
1/\nagents.
\label{eq:jm-eigentrust}
\end{equation}
The computation starts from $\boldsymbol p$ and stops when the $\ell_1$ change is at most $10^{-10}$ or after 100 iterations. The $0.15$ damping, uniform prior, and stopping rule are fixed settings of this adaptation. This adaptation applies EigenTrust's repeated trust calculation to the feedback available in our setting. It does not assume that agents provide independent ratings of one another.

\heading{TrueSkill.} TrueSkill~\citep{herbrich2007trueskill} maintains a skill estimate and its uncertainty for each agent:
\begin{equation}
\theta_i
\sim
\mathcal N(\mu_i,\sigma_i^2).
\end{equation}
After each task, agents are ranked according to decreasing $u_{i,t}$. Scores that differ by at most $10^{-12}$ are treated as tied. TrueSkill updates $\mu_i$ and $\sigma_i$ from these rankings. We use the initialization and update defaults of the Python \texttt{trueskill} implementation~\citep{lee2026trueskillsoftware}:
\begin{equation}
\mu_0=25,
\qquad
\sigma_0=25/3,
\qquad
\beta_{\rm TS}=25/6,
\qquad
\tau=25/300,
\qquad
p_{\rm draw}=0.10.
\label{eq:jm-trueskill-settings}
\end{equation}
The score
\begin{equation}
h_i^{\rm TS}
=
\mu_i-3\sigma_i
\end{equation}
favors agents with high estimated skill and low uncertainty. We convert it into a positive weight for the next task:
\begin{equation}
w_{i,t+1}
\propto
\exp
\left(
\operatorname{clip}_{[-50,50]}
\frac{
h_i^{\rm TS}-\min_jh_j^{\rm TS}
}{
\max\{1,\beta_{\rm TS}\}
}
\right).
\label{eq:jm-trueskill-weight}
\end{equation}
This baseline evaluates whether accounting for uncertainty in an agent's estimated capability improves aggregation. It does not maintain a separate record of suspicious behavior.

\heading{Task-dependent aggregation rules.} These rules implement $\aggfunc$ in \equaref{eq:overview-rep-aggregation}, corresponding to weighted aggregation in \figrref{fig:overview}. After each method determines the aggregation pool $P_t$ and weights $\{w_{i,t}\}_{i\in P_t}$, a task-dependent aggregation rule converts the selected proposals into the final answer $\hat y_t$. These rules are used by all evaluated methods. The MATH and HumanEval Pro rules are identical across methods. For GoEmotions, the seven baselines share one rule, while \rep uses the alternative rule defined below. The task-specific scoring coefficients, label thresholds, and label-count cap are fixed choices of these aggregation implementations, not benchmark-prescribed defaults.

Let $V_t\subseteq P_t$ contain the agents whose proposals provide valid, nonempty candidate answers, and let $W_t=\sum_{i\in V_t}w_{i,t}$. Proposals without a valid candidate answer receive no aggregation weight. If $V_t$ is empty, the method returns an empty answer, which is evaluated as incorrect.

For MATH, we group candidate answers according to mathematical equivalence and sum the weights within each group. The group with the largest total weight determines the final answer. This comparison uses only the submitted candidate answers and does not use the reference answer.

For HumanEval Pro, the aggregation rule selects the candidate program $i\in V_t$ with the highest score:
\begin{equation}
2p_i^{\rm example}
+0.9\frac{\sum_{j\in V_t:\,F_j=F_i}w_{j,t}}{W_t}
+0.35\frac{w_{i,t}}{W_t}
+0.25s_i^{\rm static},
\label{eq:jm-code-aggregation}
\end{equation}
where $p_i^{\rm example}$ measures whether the program passes the examples included in the task, $F_i$ identifies programs with the same normalized abstract syntax tree, and $s_i^{\rm static}$ checks whether the program is syntactically valid and contains the required entry point. The hidden evaluation tests are not used during aggregation.

For GoEmotions, the seven baselines distribute each proposal's weight uniformly across its predicted label set $Y_i$:
\begin{equation}
A_\ell=\sum_{i\in V_t}w_{i,t}\frac{\mathbf{1}[\ell\in Y_i]}{|Y_i|}.
\end{equation}
They return at most five labels satisfying
\begin{equation}
A_\ell\ge\max\left\{0.45\max_h A_h,\;0.18W_t\right\},
\end{equation}
and return the highest-scoring label if no label satisfies this condition. In contrast, \rep selects one of the label sets proposed by the agents. It chooses the set $Y$ that maximizes
\begin{equation}
0.50\frac{\sum_{i\in V_t:Y_i=Y}w_{i,t}}{W_t}
+0.40\frac{\sum_{i\in V_t}w_{i,t}\,\mathrm{F1}(Y,Y_i)}{W_t}
+0.10\frac{|\{g(i):i\in V_t,\ Y_i=Y\}|}{|\{g(i):i\in V_t\}|},
\label{eq:jm-set-aggregation}
\end{equation}
where $g(i)$ denotes the clone group of agent $i$. This rule considers the total weight supporting $Y$, its agreement with the other proposed label sets, and the number of clone groups that propose it. Therefore, the GoEmotions experiments compare the complete aggregation procedures rather than isolating the effect of reputation weights alone.

\section{Deferred Design Details of \rep}
\label{app:design}

The following details are organized by the three modules in \figrref{fig:overview} and \secrref{sec:design}.

Unless attributed to a cited method or checkpoint calibration, the numerical settings below are fixed choices of our implementation.

\subsection{Response Analyzer}
\label{app:design:analyzer}

\subsubsection{Public-evidence Score} \label{app:design:evidence}
This is the public-support score used by the Response Analyzer in \secrref{sec:design}. Each candidate receives a score computed from the cohort alone, with no gold labels and no hidden tests.
\begin{equation}
p^{pub}_{i,t} = \eta_{i,t}\bigl(
  \beta_1 \pi_{i,t} + \beta_2 \gamma_{i,t} + \beta_3 \lambda_{i,t}
  + \beta_4 s^{self}_{i,t} + \beta_5 (1-\varepsilon_t)\bigr),
\label{eq:public-support}
\end{equation}
Here $\pi$ is cohort-agreement probability, $\gamma$ cross-model agreement, and $\lambda$ leave-one-group-out plurality agreement. The remaining terms are same-model self-consistency $s^{self}$ and normalized answer-distribution entropy $\varepsilon_t$. We fix the implementation weights at $\beta=(0.45,0.25,0.15,0.10,0.05)$, and $\eta_{i,t}$ is the parse-validity indicator.
We compute all five statistics after grouping identities by underlying model, so replicas of one model count as one evidence source.

\subsubsection{Local feedback models} \label{app:design:feedback}
For the semantic verifier in \secrref{sec:design}, we fine-tune \texttt{Qwen2.5-1.5B-Instruct} as a binary correctness classifier~\citep{qwen2025qwen25}.
Its input contains the task with gold fields removed, the candidate response, parse validity, answer-group sizes, same-model consistency, cohort agreement, and normalized cohort entropy. Agent and model identifiers, attack names, Byzantine labels, gold answers, and hidden tests are excluded.
We use 4-bit LoRA~\citep{hu2022lora,dettmers2023qlora} ($r=16$, $\alpha=32$) with class-weighted focal cross-entropy~\citep{lin2017focal} and a soft-ECE regularizer~\citep{karandikar2021soft}, followed by temperature scaling~\citep{guo2017calibration}. The rank and scaling factor are fixed fine-tuning settings of our verifier, rather than defaults prescribed by these methods.
\begin{equation}
\mathcal{L}_{\mathrm{ver}} =
\frac{1}{N}\sum_j (1-p_{y_j})^\gamma \mathrm{CE}(z_j, y_j)
+ \lambda_{\mathrm{cal}} \widetilde{\mathrm{ECE}},
\label{eq:verifier-loss}
\end{equation}
Splits are made by canonical task, which prevents attack variants of one task from crossing partitions. The calibrated output is $p^{sem}_{i,t}$.
The attack families used to train the local feedback models are \textsc{Rand}, \textsc{Strong}, \textsc{Div}, \textsc{OnOff}, and \textsc{Adapt}; \textsc{Arith} and \textsc{Bound} are excluded from that training and evaluated in \appref{app:feedback-generalization}.

\heading{Behavior probe.} This probe provides the behavior-model signal for risk detection in \figrref{fig:overview}. Correctness alone may miss coordination or build-then-attack behavior.
We train an $11\!\rightarrow\!16\!\rightarrow\!8\!\rightarrow\!1$ MLP on eleven label-free features: parse failure, semantic incorrectness, self-inconsistency, cohort disagreement, cohort entropy, confidence gap, response-length outlier, leave-one-out plurality flip, cross-round answer change, and two lexical indicators for instruction-override and consensus-spoofing phrasing.
Its target is an active payload, never a static Byzantine identity. After standardization and temperature scaling it outputs $r^{atk}_{i,t}$. The MLP layer widths are fixed architectural choices; its weights are learned from the active-payload labels described above.

\subsubsection{Computing quality, risk, and blocking signals}
\label{app:design:fusion}

This subsection details the quality $q_{i,t}$, risk $r^{atk}_{i,t}$, validity $\eta_{i,t}$, and blocking $d_{i,t}$ signals produced by the Response Analyzer in \secrref{sec:design}. The Reputation-Aware Aggregator uses these signals to select agents and determine their weights. They are also passed to the State Updater after the final answer is produced.

\heading{Proposal quality.} The Response Analyzer combines the public-support score $p^{pub}_{i,t}$ with the correctness probability $p^{sem}_{i,t}$ produced by the semantic verifier:
\begin{equation}
s_{i,t}
=
p^{pub}_{i,t}
+
\omega_{i,t}
\left(
p^{sem}_{i,t}-p^{pub}_{i,t}
\right),
\qquad
\omega_{i,t}
=
\omega_0
\left(
1-(1-\kappa)
\mathbf{1}
\left[
p^{pub}_{i,t}\ge\theta_{\mathrm{res}}
\wedge
p^{sem}_{i,t}<p^{pub}_{i,t}
\right]
\right).
\label{eq:fusion-weight}
\end{equation}
The semantic verifier normally receives weight $\omega_0$. This weight is reduced when the public-support score is high but the semantic verifier gives a lower score. This prevents one low semantic-verifier score from overriding strong support from the information available on the current task. The two sources of evidence therefore contribute to quality estimation without either acting as an unconditional veto. The Response Analyzer then accounts for whether the proposal follows the required output format:
\begin{equation}
q_{i,t}
=
s_{i,t}
\left(
0.75+0.25c^{fmt}_{i,t}
\right),
\qquad
\neg\eta_{i,t}
\Rightarrow
q_{i,t}=0.
\label{eq:proposal-quality}
\end{equation}
Here, $c^{fmt}_{i,t}\in[0,1]$ measures whether the proposal follows the expected output format. As defined in the main body, $\eta_{i,t}=1$ means that a candidate answer can be extracted from the proposal. If no candidate answer can be extracted, its quality is set to zero.

\heading{Risk and blocking.} The behavior model produces the risk score $r^{atk}_{i,t}$. It also checks for an instruction override, a false claim of agreement with other proposals, disagreement with the other proposals, and an unexpected change from the agent's earlier response. Let $j_{i,t}\in[0,1]$ be the larger of the instruction-override and false-agreement scores. Let $g_{i,t}$ measure agreement with the other proposals, and let $s^{self}_{i,t}$ measure same-model self-consistency, as in \equaref{eq:public-support}. We define
\begin{equation}
b^{res}_{i,t}
=
\mathbf{1}
\left[
\eta_{i,t}
\wedge
j_{i,t}=0
\wedge
g_{i,t}\ge0.65
\wedge
s^{self}_{i,t}\ge0.65
\right].
\end{equation}
When $b^{res}_{i,t}=1$, the proposal can be parsed, contains no detected instruction override or false claim of agreement, and satisfies both the cohort-agreement and same-model consistency thresholds. This condition prevents such a proposal from being flagged only because one of its other scores is low.

The Response Analyzer identifies proposals that satisfy at least one of the following conditions:
\begin{equation}
\begin{aligned}
e^{hard}_{i,t}
&=
\mathbf{1}
\left[
r^{atk}_{i,t}\ge0.85
\wedge
j_{i,t}>0
\right],\\
o_{i,t}
&=
\mathbf{1}
\left[
p^{sem}_{i,t}\le0.20
\wedge
r^{atk}_{i,t}\ge\max\{0.50,0.75\theta_p\}
\wedge
1-g_{i,t}\ge0.50
\wedge
\neg b^{res}_{i,t}
\right],\\
C_{i,t}
&=
\mathbf{1}[\neg\eta_{i,t}]
\vee
e^{hard}_{i,t}
\vee
o_{i,t}
\vee
\mathbf{1}
\left[
p^{sem}_{i,t}<\theta_s
\wedge
r^{atk}_{i,t}\ge\theta_p
\wedge
\neg b^{res}_{i,t}
\right].
\end{aligned}
\label{eq:semantic-outlier}
\end{equation}
The first condition detects a proposal with both high behavior-model risk and a detected instruction override or false claim of agreement. The second detects a proposal with low estimated correctness, sufficiently high behavior-model risk, and substantial disagreement with the other proposals. The final condition combines a low correctness probability with a high behavior-model risk. We use checkpoint-calibrated thresholds $\theta_s=0.158765$ and $\theta_p=0.676139$, fixed across the reported conditions. For a parse-valid proposal, low estimated correctness alone is therefore insufficient to trigger a detector flag.

To select which proposals to flag, the Response Analyzer combines the behavior-model score, the semantic-verifier score, $j_{i,t}$, and validity:
\begin{equation}
a_{i,t}
=
\operatorname{clip}_{[0,1]}
\left[
0.55r^{atk}_{i,t}
+
0.30(1-p^{sem}_{i,t})
+
0.10j_{i,t}
+
0.05(1-\eta_{i,t})
\right].
\end{equation}
Among the proposals satisfying $C_{i,t}=1$, let $\mathcal F_t$ contain up to $\lceil0.30\nagents\rceil$ agents with the highest $a_{i,t}$. Ties are resolved by agent identity. The detector flag $f_{i,t}$ and the blocking signal $d_{i,t}$ are
\begin{equation}
f_{i,t}
=
\mathbf{1}[i\in\mathcal F_t],
\qquad
d_{i,t}
=
\mathbf{1}[\neg\eta_{i,t}]
\vee
f_{i,t},
\qquad
\sum_i f_{i,t}
\le
\lceil0.30\nagents\rceil.
\label{eq:behavior-alarm}
\end{equation}
The limit applies only to proposals flagged by the detector. The $0.30$ cap matches the configured corruption fraction $\nfaulty/\nagents=3/10$; the remaining fusion and rescue constants are fixed implementation settings. A proposal from which no candidate answer can be extracted is always blocked, so such proposals are not subject to this limit. The State Updater records
\begin{equation}
\widehat a_{i,t}
=
\max\{a_{i,t},0.90d_{i,t}\}
\end{equation}
after the final answer is produced.

\subsection{Reputation-Aware Aggregator}
\label{app:design:aggregator}

\subsubsection{Adjusting current weights and selecting agents}
\label{app:design:weighting}

This subsection details reputation calibration and group-aware selection in the Reputation-Aware Aggregator shown in \figrref{fig:overview}. It uses the signals defined above together with the historical weight $w^{hist}_{i,t}$ to select the pool $P_t$ and determine the aggregation weights $\{w_{i,t}\}_{i\in P_t}$.

The Reputation-Aware Aggregator first converts the current and historical risk records into a value used to reduce the current weight:
\begin{equation}
r^{gate}_{i,t}
=
\begin{cases}
\max\{a_{i,t},0.90\},
&
d_{i,t}=1,\\
\max\{0.25a_{i,t},0.50I_{i,t-1}\},
&
d_{i,t}=0.
\end{cases}
\label{eq:gate-risk}
\end{equation}
A blocked proposal receives a value of at least $0.90$. Otherwise, this value combines the risk of the current proposal with the suspicious-behavior record $I_{i,t-1}$ from previous tasks. Current evidence can thus reduce influence before the historical reputation is updated.

The adjusted weight before applying the clone-group constraint is
\begin{equation}
\begin{aligned}
\widetilde w_{i,t}
&=
\operatorname{clip}_{[\epsilon,10^6]}
\left[
w^{hist}_{i,t}
\exp
\left(
3(q_{i,t}-q_t^{med})
-
8r^{gate}_{i,t}
\right)
m_{i,t}
\right],\\
m_{i,t}
&=
0.01^{d_{i,t}}
\cdot
0.02^{Q_{i,t}}
\cdot
\epsilon^{1-\eta_{i,t}}.
\end{aligned}
\label{eq:current-weight}
\end{equation}
Here, $q_t^{med}$ is the median quality among proposals from which a candidate answer can be extracted, and it is zero if no such proposal exists. We use $\epsilon=10^{-12}$. The floor and clipping bound are numerical safeguards; the quality/risk coefficients, penalty multipliers, and clone-group constraints below are fixed aggregation settings. The indicator $Q_{i,t}=1$ means that previous tasks identified the agent as suspicious or that its temporary restriction remains active. The formula increases the weight of proposals with quality above the current-task median and reduces the weight of proposals with high risk. The multiplier $m_{i,t}$ further reduces the weight of a blocked agent, an agent with a previous suspicious-behavior record, or a proposal from which no candidate answer can be extracted.

The Reputation-Aware Aggregator then limits the combined influence of agents in the same clone group. For a clone group $G$ containing $m=|G|$ agents, define
\begin{equation}
M_G
=
\sum_{i\in G}\widetilde w_{i,t},
\qquad
p_G
=
\max_{i\in G}\widetilde w_{i,t}.
\end{equation}
The total weight of the group is limited to
\begin{equation}
\min
\left\{
M_G,
p_G
\left[
1+0.15(\sqrt m-1)
\right]
\right\},
\end{equation}
and the weights of its members are reduced proportionally when this limit is exceeded. For example, three agents with equal weights from the same clone group receive at most $1.11$ times the weight of one member, rather than three times that weight.

Finally, the Reputation-Aware Aggregator selects $k$ agents for $P_t$. It first attempts to include agents from at least three clone groups and no more than three agents from one clone group. These requirements are relaxed only when necessary to fill all $k$ positions. Agents blocked on the current task and agents whose temporary restrictions remain active are considered after the other agents. After selecting $P_t$, the Reputation-Aware Aggregator applies the quality adjustment and clone-group limit again within the selected pool. It then normalizes the resulting values to obtain $\{w_{i,t}\}_{i\in P_t}$. The selected proposals and weights are then combined using the task-dependent aggregation rules in \appref{app:baseline}.

\subsection{State Updater}
\label{app:design:updater}

\subsubsection{Updating records of suspicious behavior}
\label{app:design:identity}

This subsection details the behavioral-risk update in the State Updater shown in \figrref{fig:overview}. After the Reputation-Aware Aggregator produces $\hat y_t$, the State Updater records the quality and risk of each valid proposal. These records reduce an agent's historical weight on subsequent tasks after suspicious behavior and allow its influence to recover after sufficiently many proposals without a detector flag. The decay rates, risk thresholds, three-task cooldown, and recovery constants in this subsection are fixed state-update settings.

\heading{Responses without an extracted candidate answer.} If $\eta_{i,t}=0$, the proposal is excluded from the current aggregation. The State Updater does not treat the parse failure itself as evidence of a malicious proposal. The number of valid observations $n_i$, the quality records, and the suspicious-behavior records $D_i$, $K_i$, $E_i$, and $I_i$ remain unchanged. The counter $h_i$ for the temporary restriction is reduced by one:
\begin{equation}
h_i
\leftarrow
\max(0,h_i-1).
\end{equation}
Thus, a parse failure does not create a new quality observation and does not remove a previously recorded risk.

\heading{Updating suspicious-behavior records.} The State Updater maintains $E_i$ as a recent record of detector results and $K_i$ as a record that gives additional weight to detector flags and other high-risk observations. Before applying the recovery rule below, they are updated as
\begin{equation}
\begin{aligned}
E'_{i,t}
&=
0.65E_{i,t-1}
+
0.35
\left[
d_{i,t}\widehat a_{i,t}
+
(1-d_{i,t})0.30a_{i,t}
\right],\\
K'_{i,t}
&=
\operatorname{clip}_{[0,1]}
\left[
0.90K_{i,t-1}
+
0.55d_{i,t}
+
0.15(1-d_{i,t})
r^{inst}_{i,t}
\mathbf{1}[r^{inst}_{i,t}\ge0.65]
\right].
\end{aligned}
\label{eq:identity-risk}
\end{equation}
When a valid proposal is flagged, the temporary-restriction counter is set to at least three tasks:
\begin{equation}
h_{i,t}
=
\max
\left\{
\max(0,h_{i,t-1}-1),
3
\right\}.
\end{equation}
Otherwise, the counter decreases by one. After applying the recovery rule, the combined suspicious-behavior record is
\begin{equation}
I_{i,t}
=
\max
\left\{
E_{i,t},
K_{i,t},
0.75D_{i,t}
\mathbf{1}[\text{the agent has previously been flagged}]
\right\}.
\end{equation}
The agent is treated as previously suspicious when $I_{i,t}\ge0.55$ or $h_{i,t}>0$. This sets $Q_{i,t+1}=1$ when the Reputation-Aware Aggregator processes the next task.

\heading{Reducing the effect of earlier suspicious behavior.} A valid proposal contributes to recovery when it is not flagged and its quality satisfies
\begin{equation}
q_{i,t}
\ge
q_t^{med}-0.05.
\end{equation}
The State Updater records the number of consecutive proposals satisfying these conditions. A valid proposal that does not satisfy them resets this count, while a proposal with $\eta_{i,t}=0$ leaves it unchanged. Once the count reaches $\nu=3$, each subsequent qualifying proposal reduces the suspicious-behavior records:
\begin{equation}
E_{i,t}
=
0.75E'_{i,t},
\qquad
K_{i,t}
=
0.80K'_{i,t},
\qquad
D_{i,t}
=
0.90D'_{i,t}.
\end{equation}
Otherwise, the primed values remain unchanged. The time required for an agent to recover therefore depends on the quality and risk of its subsequent proposals. The additional decay is thus tied to repeated qualifying proposals, not just elapsed time.

\subsubsection{Computing the historical weight}
\label{app:design:histweight}

This subsection details the capability update and historical-weight computation of the State Updater in \secrref{sec:design}. The records from tasks up to $t-1$ determine the historical weight $w^{hist}_{i,t}$ used by the Reputation-Aware Aggregator on task $t$. The weight increases with the agent's long-term and recent quality. It decreases when the agent's recent performance falls below its earlier performance or when suspicious behavior has been recorded. The prior means and strength, EMA rates, historical-weight coefficients, and drift/debt constants below are likewise fixed implementation settings.

The State Updater first computes
\begin{equation}
\begin{aligned}
L_{i,t-1}
={}&
6\log(\bar e_{i,t-1}+0.03)
+
3\log(\bar q_{i,t-1}+0.03)\\
&+
1.5\log
\operatorname{clip}_{[0.05,1]}
\left(
\frac{q^{short}_{i,t-1}}{\bar q_{i,t-1}}
\right)
+
2\log
\operatorname{clip}_{[0.08,1]}
\left(
\frac{e^{short}_{i,t-1}}{\bar e_{i,t-1}}
\right)\\
&-
6r^{temp}_{i,t-1}
-
5\mathbf{1}[h_{i,t-1}>0],
\\
w^{hist}_{i,t}
={}&
\max
\left\{
\epsilon,
\exp
\left(
L_{i,t-1}
-
\max_jL_{j,t-1}
\right)
\right\}.
\end{aligned}
\label{eq:hist-weight}
\end{equation}
Here, $\bar q_i$ records long-term proposal quality, $q^{short}_i$ gives more weight to recent proposal quality, $\bar e_i$ records the long-term correctness probability produced by the semantic verifier, and $e^{short}_i$ gives more weight to recent correctness probabilities. The value $r^{temp}_i$ is the largest recorded indication of suspicious behavior or a substantial performance decrease. The counter $h_i$ records whether a temporary restriction remains active. Taking the largest risk value ensures that a high risk recorded from one source is not cancelled by lower values from the other records.

\heading{Updating long-term and recent quality.} We initialize
\begin{equation}
\bar q_i=q^{short}_i=0.30,
\qquad
\bar e_i=e^{short}_i=0.18.
\end{equation}
For a valid proposal, define
\begin{equation}
u^{cap}_{i,t}
=
\begin{cases}
p^{sem}_{i,t},
&
\text{if the semantic-verifier score is available},\\
p^{pub}_{i,t},
&
\text{otherwise}.
\end{cases}
\end{equation}
The long-term records are updated as
\begin{equation}
\bar q_{i,t}
=
\frac{
(12+n_{i,t-1})\bar q_{i,t-1}
+
q_{i,t}
}{
13+n_{i,t-1}
},
\qquad
\bar e_{i,t}
=
\frac{
(12+n_{i,t-1})\bar e_{i,t-1}
+
u^{cap}_{i,t}
}{
13+n_{i,t-1}
}.
\label{eq:beta-prior}
\end{equation}
The count $n_i$ increases once for each valid proposal. The recent records are updated as
\begin{equation}
\begin{aligned}
q^{short}_{i,t}
&=
(1-\alpha_{i,t})
q^{short}_{i,t-1}
+
\alpha_{i,t}q_{i,t},
\qquad
\alpha_{i,t}
=
\begin{cases}
0.30,
&
q_{i,t}<q^{short}_{i,t-1},\\
0.08,
&
\text{otherwise},
\end{cases}\\
e^{short}_{i,t}
&=
0.80e^{short}_{i,t-1}
+
0.20u^{cap}_{i,t}.
\end{aligned}
\label{eq:asym-ema}
\end{equation}
A decrease in proposal quality therefore changes $q^{short}_i$ faster than an improvement of the same size. We use $\omega_0=0.75$, $\kappa=0.85$, and $\theta_{\mathrm{res}}=0.80$ in \equaref{eq:fusion-weight}. When the semantic-verifier score is unavailable, its weight in \equaref{eq:fusion-weight} is set to zero.

\heading{Recording substantial performance decreases.} For a valid proposal, define its decrease below the current-task median as
\begin{equation}
g^{gap}_{i,t}
=
\max
\left\{
0,
q_t^{med}-q_{i,t}-0.05
\right\}.
\end{equation}
The State Updater maintains a recent average of this value:
\begin{equation}
G_{i,t}
=
0.75G_{i,t-1}
+
0.25g^{gap}_{i,t}.
\end{equation}
It compares this recent average with the agent's earlier average $B_{i,t-1}$:
\begin{equation}
r^{chg}_{i,t}
=
\operatorname{clip}_{[0,1]}
\left[
\frac{
G_{i,t}-B_{i,t-1}-0.025
}{
0.10
}
\right].
\end{equation}
The earlier average is updated only when the proposal is not blocked and the increase is small:
\begin{equation}
B_{i,t}
=
0.995B_{i,t-1}
+
0.005g^{gap}_{i,t}
\end{equation}
when $r^{chg}_{i,t}<0.25$ and $d_{i,t}=0$. Otherwise, $B_{i,t}=B_{i,t-1}$.

The State Updater also measures low absolute quality:
\begin{equation}
r^{abs}_{i,t}
=
\operatorname{clip}_{[0,1]}
\left[
\frac{0.50-q_{i,t}}{0.50}
\right],
\qquad
r^{inst}_{i,t}
=
\max
\left\{
r^{chg}_{i,t},
r^{abs}_{i,t},
\widehat a_{i,t}
\right\}.
\end{equation}
It then updates $D_i$, which records recent evidence of a substantial performance decrease:
\begin{equation}
\begin{aligned}
\widetilde D_{i,t}
&=
0.75D_{i,t-1}
+
0.25
\max
\left\{
r^{chg}_{i,t},
0.70r^{abs}_{i,t},
\widehat a_{i,t}
\right\},\\
D'_{i,t}
&=
\begin{cases}
\max\{\widetilde D_{i,t},0.90\},
&
d_{i,t}=1,\\
\max\{\widetilde D_{i,t},0.54r^{inst}_{i,t}\},
&
d_{i,t}=0
\ \wedge\
r^{inst}_{i,t}\ge0.65,\\
\widetilde D_{i,t},
&
\text{otherwise}.
\end{cases}
\end{aligned}
\label{eq:conditional-debt}
\end{equation}
Finally, $r^{temp}_i$ is the maximum of $D_i$, $E_i$, $I_i$, $r^{inst}_i$, and the decrease of the long-term quality records from their previous high values. To measure this last decrease, the State Updater maintains
\begin{equation}
H^x_{i,t}
=
\max
\left\{
\bar x_{i,t},
0.9999H^x_{i,t-1}
\right\},
\qquad
x\in\{q,e\}.
\end{equation}
It compares $\bar q_i$ and $\bar e_i$ with $H^q_i$ and $H^e_i$, respectively. A large decrease in either value increases $r^{temp}_i$. Proposals with $\eta_{i,t}=0$ do not update these records.

\section{Detailed Experiment Setups} \label{app:experiment}
\subsection{Datasets}
This subsection gives the sampling details for the benchmarks in \secrref{sec:evaluation:setup}.

We select 100 GoEmotions validation items without changing their order in the original dataset. For MATH, we shuffle the level-4/5 test items using seed 2027 and select 100 tasks. We use the same procedure to select 100 HumanEval Pro tasks from the split labeled \texttt{train} in its upstream release. These HumanEval Pro tasks are used only for evaluation. The 100-task sample size and seed 2027 are choices of our evaluation protocol, not defaults of the source benchmarks.

\subsection{Agent Pools and Configurations}\label{sec:pool}

\heading{Model rankings and clone groups.} We rank the seven LLMs separately for each dataset using their standalone task scores. Each model has three standalone evaluations on the fixed evaluation tasks. For GoEmotions, the three evaluations use different personality prompts. For MATH, we retain the ranking used when constructing the agent pools and do not recompute it using the final answer-equivalence rule. From strongest to weakest, the rankings are Qwen-Max, Qwen-Flash, GLM, Qwen-Plus, DeepSeek, Kimi, and MiniMax for GoEmotions; Kimi, MiniMax, Qwen-Plus, Qwen-Max, GLM, Qwen-Flash, and DeepSeek for MATH; and Qwen-Max, Qwen-Plus, GLM, Qwen-Flash, DeepSeek, Kimi, and MiniMax for HumanEval Pro.

These rankings determine the four agent compositions in \tabref{tab:design}. The all-strong composition contains agents instantiated from the four highest-ranked models, with $3$, $3$, $2$, and $2$ agents per model. The all-weak composition uses models ranked fourth through seventh, with $2$, $2$, $3$, and $3$ agents per model. The 7S+3W composition contains $3$, $2$, and $2$ agents from the three highest-ranked models and one agent from each of the three lowest-ranked models. The random composition uses a fixed selection generated with seed 2027. Agents instantiated from the same underlying model form a clone group.

The rankings also determine the corruption placements in \tabref{tab:design}. B and W compromise agents instantiated from higher- and lower-ranked models, respectively. R1 and R2 are two fixed subsets selected using seed 2027. The detailed result tables report the agents compromised in each selected condition.

\heading{Run configuration.} We execute 128 runs per dataset: 112 attacked runs from four compositions, four corruption placements, and seven attacks, and 16 clean runs from four compositions and four placements. This gives 384 runs across the three datasets. For a fixed composition, the four clean placements produce the same task scores and are counted once when computing clean averages. The semantic verifier and behavior model are shared across methods; each method's state-update parameters are fixed across experimental conditions. The OnOff and Adapt attacks begin after the first 30 tasks. This 30-task schedule and the pool, corruption, and selection sizes in \secrref{sec:evaluation:setup} are fixed controlled-evaluation settings. For Adapt, the active malicious agents are selected according to the current \rep state, and the resulting attack plan is applied to every method in the same experimental condition.

\heading{Feedback and aggregation.} This setup implements the comparison described in \secrref{sec:evaluation:setup}, using the task-dependent aggregation rules in \appref{app:baseline}. All stateful methods receive the same quality assessment from the semantic verifier after each task. \rep additionally uses this assessment in the Response Analyzer before aggregation and in the State Updater after the final answer is produced. On GoEmotions, the baselines distribute each proposal's weight among its predicted labels. \rep instead selects one of the label sets submitted by the agents based on their weights, agreement, and clone groups. Therefore, the GoEmotions results compare the complete methods, including their aggregation rules, rather than isolating only the effect of historical reputation.

\subsection{Metrics} \label{app:metric}

We define the task metrics reported in \secrref{sec:evaluation:setup} and the additional diagnostics reported in \appref{sec:additional-results}. 

\heading{F1, accuracy, and Pass@1.} These standard task metrics measure label overlap, exact correctness, and successful execution, respectively. For $N$ evaluated tasks, let $\hat y_t$ be the final aggregate and $y_t$ the reference. Recall that the sample-averaged F1 score~\citep{sokolova2009systematic} and exact-match accuracy are defined as:
\begin{equation}
\mathrm{F1}
=
\frac{1}{N}
\sum_{t=1}^{N}
\frac{2|\widehat Y_t\cap Y_t|}
{|\widehat Y_t|+|Y_t|},
\qquad
\mathrm{Acc}_{\rm set}
=
\frac{1}{N}
\sum_{t=1}^{N}
\mathbf{1}[\widehat Y_t=Y_t\ne\varnothing],
\label{eq:jm-go-metrics}
\end{equation}
where $\widehat Y_t$ and $Y_t$ denote the predicted and reference label sets after normalization. We use these metrics for the GoEmotions dataset. In addition, for MATH, we model the accuracy as $N^{-1}\sum_t\mathbf{1}[\hat y_t\equiv y_t]$, where $\equiv$ is the evaluator's answer-equivalence predicate. For HumanEval Pro, we use the Pass@k metric~\citep{chen2021evaluating} where $k=1$, denoted as $\mathrm{Pass@1}=N^{-1}\sum_t\mathbf1[\hat y_t\text{ passes all evaluation tests}]$.

\heading{$\Delta$ and RelDrop: Paired loss and retention.} These measures quantify the clean-to-attacked performance changes discussed in \secrref{sec:evaluation:results}. For method $m$ and $N$ evaluated tasks, let $s^0_{m,t}$ and $s^a_{m,t}$ denote its clean and attacked scores on task $t$. Their averages are $S^0_m=N^{-1}\sum_t s^0_{m,t}$ and $S^a_m=N^{-1}\sum_t s^a_{m,t}$. Each clean--attack pair uses the same method, tasks, seed, agent composition, and cached proposals. The clean run executes the complete protocol without attacks and maintains its own reputation scores.

We define the paired loss $\Delta_m$, relative performance drop $\mathrm{RelDrop}_m$, and retention advantage over baseline $b$, denoted by $R_b$, as:
\begin{equation}
\Delta_m
=
100(S^0_m-S^a_m),
\qquad
\mathrm{RelDrop}_m
=
100\frac{S^0_m-S^a_m}{S^0_m},
\qquad
R_b
=
\Delta_b-\Delta_{\rep}.
\label{eq:jm-paired-loss}
\end{equation}
$\Delta_m$ and $R_b$ are measured in percentage points. RelDrop is a percentage and is undefined when $S^0_m=0$. A negative loss means that the attacked run performs better than its clean counterpart. A positive $R_b$ means that \rep loses fewer points than baseline $b$ from their respective clean scores. It does not necessarily mean that \rep has a higher attacked score. The paired comparison separates attack-related performance loss from differences in clean performance.

\heading{PoolExp and PayloadHit: Current-task exposure and payload agreement.} These diagnostics examine pool selection and final aggregation in \figrref{fig:overview}; results appear in \tabref{tab:average_defense}. Let $A_t=1$ indicate that an attack is active on task $t$, let $J_t$ denote the agents whose proposals contain an attack payload, and let $P_t$ denote the aggregation pool. We define $E_t=\mathbf{1}[P_t\cap J_t\ne\varnothing]$ to indicate that at least one attack payload enters the aggregation pool. We also define $H_t=\mathbf{1}[\hat y_t\text{ matches a payload from }P_t\cap J_t]$. PoolExp and PayloadHit are defined as:
\begin{equation}
\mathrm{PoolExp}
=
\frac{\sum_t E_t}{\sum_t A_t},
\qquad
\mathrm{PayloadHit}
=
\frac{\sum_t E_tH_t}{\sum_t E_t}.
\label{eq:jm-exposure-metrics}
\end{equation}
PoolExp measures the fraction of tasks with $A_t=1$ for which at least one proposal containing an attack payload enters the aggregation pool. PayloadHit measures the fraction of tasks with $E_t=1$ for which the final output matches an attack payload in the aggregation pool. A match is determined using label-set equality for GoEmotions, answer equivalence for MATH, and equality between the normalized program structures for HumanEval Pro. PayloadHit is undefined when $\sum_t E_t=0$.

\heading{NextExcl: Exclusion from the next effective pool.} This diagnostic examines subsequent pool selection after the state update in \figrref{fig:overview}; results appear in \tabref{tab:average_defense}. Let $B_t$ denote the set of agents that actively attack on task $t$. Let $\tau(t)$ denote the first subsequent task whose pool selection can use the information collected on task $t$. With immediate updates, $\tau(t)=t+1$. We define $\mathcal{T}^{+}=\{t:B_t\ne\varnothing,\ \tau(t)\text{ is evaluated}\}$. NextExcl is defined as:
\begin{equation}
\mathrm{NextExcl}
=
\frac{1}{|\mathcal{T}^{+}|}
\sum_{t\in\mathcal{T}^{+}}
\mathbf{1}[B_t\cap P_{\tau(t)}=\varnothing].
\label{eq:jm-next-exclusion}
\end{equation}
NextExcl measures the fraction of eligible attack tasks for which all agents that attacked on task $t$ are excluded from the next pool affected by the updated reputation state. Tasks without an evaluated future pool are omitted. The metric evaluates exclusion at the task level rather than separately for each attacker.

\heading{CF-Harm: Counterfactual harm.} This diagnostic supplements the task scores in \secrref{sec:evaluation:results} by measuring how often attacks reduce per-task performance. Let $s^0_{m,t}$ and $s^a_{m,t}$ denote the clean and attacked scores of method $m$ on task $t$. Using the paired clean and attacked runs, we define CF-Harm as:
\begin{equation}
\mathrm{CF\text{-}Harm}
=
\frac{
\sum_t A_t\,
\mathbf{1}[s^0_{m,t}-s^a_{m,t}>10^{-12}]
}{
\sum_t A_t
}.
\label{eq:jm-counterfactual-harm}
\end{equation}
CF-Harm measures the fraction of active-attack tasks on which the attacked score is lower than the paired clean score. It measures how frequently an attack causes harm rather than the magnitude of that harm. For MATH and HumanEval Pro, it counts tasks that are correct in the clean run but incorrect in the attacked run. For GoEmotions, it also captures partial decreases in sample F1. Unlike the signed paired loss $\Delta$, improvements on other tasks cannot cancel these harmful cases.

\subsubsection{Averaging and comparison rules}
\label{app:average_protocol}

These rules define the averages and baseline comparisons in \tabref{tab:average_utility}, \tabref{tab:main_result_summary}, and \appref{sec:additional-results}. For a method $m$, let $c\in\{1,\ldots,4\}$ denote the agent composition, $p\in\{1,\ldots,4\}$ the corruption placement, and $a\in\{1,\ldots,7\}$ the attack. Let $S^0_{m,c}$ be the clean score for composition $c$, and let $S^{\mathrm{atk}}_{m,c,p,a}$ be the attacked score for one combination of composition, placement, and attack.

\heading{Average task scores.} The clean average counts each of the four agent compositions once. The attacked average assigns equal weight to all $4\times4\times7=112$ attack conditions:
\begin{equation}
\overline S^0_m
=
\frac{1}{4}\sum_{c=1}^{4}S^0_{m,c},
\qquad
\overline S^{\mathrm{atk}}_m
=
\frac{1}{112}
\sum_{c=1}^{4}
\sum_{p=1}^{4}
\sum_{a=1}^{7}
S^{\mathrm{atk}}_{m,c,p,a}.
\label{eq:jm-sweep-averages}
\end{equation}
The average paired loss is
\begin{equation}
\overline\Delta_m
=
\frac{100}{112}
\sum_{c=1}^{4}
\sum_{p=1}^{4}
\sum_{a=1}^{7}
\left(
S^0_{m,c}-S^{\mathrm{atk}}_{m,c,p,a}
\right).
\label{eq:jm-average-loss}
\end{equation}
For RelDrop, we first compute the relative loss of each attack condition and then average these values. Let
\begin{equation}
\mathcal C_m
=
\left\{
(c,p,a):S^0_{m,c}>0
\right\}.
\end{equation}
We define
\begin{equation}
\overline{\mathrm{RelDrop}}_m
=
\frac{100}{|\mathcal C_m|}
\sum_{(c,p,a)\in\mathcal C_m}
\frac{
S^0_{m,c}-S^{\mathrm{atk}}_{m,c,p,a}
}{
S^0_{m,c}
}.
\label{eq:jm-relative-average}
\end{equation}
Thus, RelDrop is not computed by dividing the difference between the two average scores by the average clean score.

\heading{Average defense rates.} PoolExp, PayloadHit, NextExcl, and CF-Harm are first computed separately for each attack condition. Each condition for which a metric is defined contributes equally to its reported average. If a denominator is zero, that condition is omitted from the average for that metric rather than assigned a value of zero. PoolExp, NextExcl, and CF-Harm are defined in all 112 attack conditions for every dataset and method. PayloadHit is undefined when no proposal containing an attack payload enters the aggregation pool. In the method order UMaj, URand, Single, Babylon, Eigen, Beta, TS, and \rep, the numbers of conditions with defined PayloadHit are
\begin{equation}
\begin{aligned}
\text{GoEmotions:}\quad &(112,112,112,111,112,110,109,112),\\
\text{MATH:}\quad &(112,112,112,112,112,112,110,111),\\
\text{HumanEval Pro:}\quad &(112,112,111,111,111,110,112,106).
\end{aligned}
\end{equation}

\heading{Baseline selection and strict wins.} In \tabref{tab:main_result_summary}(b)--(c), \tabref{tab:paired_preservation}, and \tabref{tab:late_attack_code}, baseline $b$ is the baseline with the highest attacked score under the same dataset, composition, corruption placement, and attack. Score ties are resolved using the smaller paired loss and then the displayed method order. In \tabref{tab:average_utility}, the best baseline is selected separately for the clean and attacked averages. The two selected baselines may therefore be different, and their scores do not define a paired loss.

A \textit{strict win} requires a method's unrounded task score to exceed the scores of all seven other methods; ties within $10^{-12}$ are excluded. Each agent composition has $4\times7=28$ attack conditions. In \tabref{tab:main_result_summary}(a), the displayed baseline is the baseline with the largest number of strict wins. These counts summarize the evaluated conditions and do not represent statistical significance across independent seeds.

\subsection{Benchmark examples}
\label{app:benchmark-examples}

These examples expand the benchmark descriptions in \secrref{sec:evaluation:setup}, illustrating the information given to agents, the expected candidate answers, and the evaluation metrics. The candidate answers below are illustrative rather than outputs recorded from our experiments. Reference answers and hidden evaluation tests are used only to evaluate the final output. They are not available to \rep when selecting or aggregating proposals.

\heading{MATH.} One level-4 Algebra problem asks for the center of the circle defined by $x^2-6x+y^2+2y=9$~\citep{hendrycks2021math}. The agents receive the equation and the request for its center. Completing the square gives $(x-3)^2+(y+1)^2=19$, so the correct answer is $(3,-1)$. A proposal containing $(3,1)$ can be successfully parsed but is mathematically incorrect. During aggregation, submitted answers are grouped according to mathematical equivalence without using the reference answer. The final output is compared with the reference answer only during evaluation. This example shows why extracting a candidate answer successfully does not imply that the answer is correct.

\heading{GoEmotions.} The validation example \texttt{eczdvun} contains the text ``Thank you. I really appreciate your response'' and is annotated with the labels $\{\textit{admiration},\textit{gratitude}\}$ in the released dataset~\citep{demszky2020goemotions}. The agents receive the text and the available emotion labels, and each candidate answer is a set of labels. Predicting only $\{\textit{gratitude}\}$ gives a sample F1 score of $2/3$ but an exact-match accuracy of zero. Predicting both reference labels gives a score of one under both metrics. This example illustrates that sample F1 gives partial credit for overlapping labels, while exact-match accuracy requires the complete reference label set.

\heading{HumanEval Pro.} In the released example with ID 0, the function \texttt{has\_close\_elements(numbers, threshold)} checks whether a list contains two numbers whose distance is strictly smaller than the threshold. The extended function \texttt{find\_close\_elements\_lists(list\_of\_lists, threshold)} returns the indices of the lists satisfying this condition~\citep{yu2025humanevalpro}. For example, the input $[[1.0,2.0,3.0],[1.0,2.8,3.0,4.0,5.0,2.0]]$ with threshold $0.5$ produces $[1]$. An agent must submit Python code implementing the required functions rather than only the output for this example. The task specification and public examples can be used during aggregation, while Pass@1 is determined by whether the selected program passes all evaluation tests. Replacing the strict comparison with a non-strict comparison may preserve the public-example output but fail when two numbers differ by exactly the threshold, illustrating a Boundary Value attack.

\section{Additional Evaluation Results}\label{sec:additional-results}

\heading{Average task performance.} \tabref{tab:average_utility_full} expands \tabref{tab:average_utility} by reporting all eight methods together with their clean scores, attacked scores, paired losses $\Delta$, and relative performance drops (RelDrop). We define $\Delta$ and RelDrop in \appref{app:metric} and describe the averaging rules in \appref{app:average_protocol}. The shaded cells are the values summarized in the main body.

\rep achieves the highest attacked average on three of the four task metrics. The exception is GoEmotions sample F1, where \rep achieves 33.94\% and Beta achieves 34.87\%. On both GoEmotions metrics, \rep has the smallest $\Delta$ and RelDrop, indicating the lowest performance degradation under attacks. On MATH, \rep achieves the highest attacked accuracy of 61.95\%, compared with 54.37\% for Beta. However, its paired loss is 4.80 percentage points, while Beta's paired loss is 0.88 percentage points. These results are not contradictory because the attacked score measures performance after attacks, while paired loss measures the change from each method's own clean score. On HumanEval Pro, \rep achieves the highest attacked Pass@1 of 78.15\%. EigenTrust has the smallest paired loss because its attacked average is slightly higher than its clean average.

\begin{table}[!htb]
\centering
\caption{Average performance for all eight methods. Clean scores are averaged over the four compositions. Attacked scores, paired losses, and relative drops are the average of 112 attack conditions. Scores and RelDrop are percentages. $\Delta$ is in percentage points. Shaded cells appear in \tabref{tab:average_utility}; bold marks the best value in each row.}
\label{tab:average_utility_full}
\footnotesize
\setlength{\tabcolsep}{3pt}
\begin{tabularx}{\linewidth}{@{}>{\raggedright\arraybackslash}p{0.22\linewidth}*{8}{>{\raggedleft\arraybackslash}X}@{}}
\toprule
Metric / setting & UMaj & URand & Single & Babylon & Eigen & Beta & TS & \rep \\
\midrule
\multicolumn{9}{@{}l}{\textit{GoEmotions / exact-match accuracy}} \\
Clean$\uparrow$ & 18.50 & 17.50 & 20.25 & 20.50 & 17.75 & \cellcolor{blue!10}21.50 & 20.50 & \cellcolor{blue!10}\textbf{21.75} \\
Attacked$\uparrow$ & 10.71 & 11.04 & 19.04 & 18.05 & 12.42 & \cellcolor{blue!10}19.59 & 17.79 & \cellcolor{blue!10}\textbf{21.16} \\
$\Delta\downarrow$ & 7.79 & 6.46 & 1.21 & 2.45 & 5.33 & 1.91 & 2.71 & \textbf{0.59} \\
RelDrop$\downarrow$ & 42.64 & 37.30 & 6.41 & 12.93 & 30.33 & 9.25 & 13.54 & \textbf{2.52} \\
\midrule
\multicolumn{9}{@{}l}{\textit{GoEmotions / sample F1}} \\
Clean$\uparrow$ & \cellcolor{blue!10}\textbf{36.42} & 33.63 & 35.81 & 35.64 & 35.32 & 35.89 & 35.56 & \cellcolor{blue!10}34.42 \\
Attacked$\uparrow$ & 27.89 & 27.76 & 34.14 & 33.72 & 28.53 & \cellcolor{blue!10}\textbf{34.87} & 33.61 & \cellcolor{blue!10}33.94 \\
$\Delta\downarrow$ & 8.53 & 5.88 & 1.67 & 1.92 & 6.79 & 1.02 & 1.95 & \textbf{0.48} \\
RelDrop$\downarrow$ & 23.18 & 17.48 & 4.62 & 5.47 & 19.04 & 2.94 & 5.42 & \textbf{1.40} \\
\midrule
\multicolumn{9}{@{}l}{\textit{MATH / mathematical-equivalence accuracy}} \\
Clean$\uparrow$ & \cellcolor{blue!10}59.25 & 56.25 & 54.75 & 54.25 & 56.00 & 55.25 & 51.75 & \cellcolor{blue!10}\textbf{66.75} \\
Attacked$\uparrow$ & 43.75 & 43.48 & 53.85 & 52.54 & 42.01 & \cellcolor{blue!10}54.37 & 48.56 & \cellcolor{blue!10}\textbf{61.95} \\
$\Delta\downarrow$ & 15.50 & 12.77 & 0.90 & 1.71 & 13.99 & \textbf{0.88} & 3.19 & 4.80 \\
RelDrop$\downarrow$ & 26.15 & 21.95 & 2.13 & 3.53 & 25.11 & \textbf{1.23} & 5.89 & 6.69 \\
\midrule
\multicolumn{9}{@{}l}{\textit{HumanEval Pro / Pass@1}} \\
Clean$\uparrow$ & \cellcolor{blue!10}\textbf{79.50} & 78.75 & 77.50 & 77.00 & 76.50 & 76.75 & 77.50 & \cellcolor{blue!10}77.75 \\
Attacked$\uparrow$ & 51.75 & 56.13 & 74.69 & 72.20 & \cellcolor{blue!10}77.92 & 73.87 & 71.01 & \cellcolor{blue!10}\textbf{78.15} \\
$\Delta\downarrow$ & 27.75 & 22.62 & 2.81 & 4.80 & \textbf{-1.42} & 2.88 & 6.49 & -0.40 \\
RelDrop$\downarrow$ & 34.90 & 28.74 & 3.63 & 6.25 & \textbf{-1.89} & 3.75 & 8.41 & -0.53 \\
\bottomrule
\end{tabularx}
\end{table}

\heading{Defense effectiveness.} These diagnostics supplement the task scores in \secrref{sec:evaluation:results} by examining how attacks affect selection and aggregation in \figrref{fig:overview}. \tabref{tab:average_defense} evaluates how the approaches prevent attack payloads from entering the aggregation pool, exclude active attackers from subsequent pools, and limit their effects on the final output over the 112 attacked conditions for each dataset. Each rate is computed separately for each condition and then averaged over the conditions with a nonzero denominator. The results show that \rep's main strength is limiting the influence of attack payloads that enter the aggregation pool, rather than always excluding them. Specifically, \rep achieves the lowest PayloadHit on all three datasets: 18.22\% on GoEmotions, 26.07\% on MATH, and 0.05\% on HumanEval Pro. It also achieves the lowest CF-Harm on GoEmotions and a near-lowest value on HumanEval Pro. Beta achieves a lower CF-Harm on MATH. These results are consistent with \rep's design: the Reputation-Aware Aggregator adjusts weights using current evidence from the Response Analyzer, so a malicious proposal can have limited influence even when its agent remains in the pool.

\begin{table}[!htb]
\centering
\caption{Average defense rates of all experiments per dataset.}
\label{tab:average_defense}
\footnotesize
\setlength{\tabcolsep}{3pt}
\begin{tabularx}{\linewidth}{@{}>{\raggedright\arraybackslash}p{0.22\linewidth}*{8}{>{\raggedleft\arraybackslash}X}@{}}
\toprule
Metric / setting & UMaj & URand & Single & Babylon & Eigen & Beta & TS & \rep \\
\midrule
\multicolumn{9}{@{}l}{\textit{GoEmotions}} \\
PoolExp$\downarrow$ & 100.00 & 95.41 & 53.72 & \textbf{46.35} & 95.49 & 47.43 & 53.20 & 81.79 \\
NextExcl$\uparrow$ & 0.00 & 4.78 & 49.84 & \textbf{55.26} & 4.71 & 53.64 & 47.93 & 19.42 \\
PayloadHit$\downarrow$ & 27.93 & 24.62 & 19.55 & 22.48 & 25.22 & 19.83 & 20.69 & \textbf{18.22} \\
CF-Harm$\downarrow$ & 27.84 & 21.02 & 9.10 & 9.64 & 22.67 & 7.89 & 10.51 & \textbf{5.82} \\
\midrule
\multicolumn{9}{@{}l}{\textit{MATH}} \\
PoolExp$\downarrow$ & 100.00 & 95.21 & \textbf{29.48} & 30.33 & 94.77 & 31.23 & 55.41 & 50.74 \\
NextExcl$\uparrow$ & 0.00 & 5.00 & \textbf{74.47} & 71.83 & 5.74 & 70.06 & 46.68 & 52.64 \\
PayloadHit$\downarrow$ & 57.43 & 54.48 & 42.04 & 52.25 & 56.69 & 41.10 & 41.64 & \textbf{26.07} \\
CF-Harm$\downarrow$ & 19.92 & 16.65 & 6.08 & 7.19 & 18.40 & \textbf{5.76} & 8.02 & 8.92 \\
\midrule
\multicolumn{9}{@{}l}{\textit{HumanEval Pro}} \\
PoolExp$\downarrow$ & 100.00 & 95.95 & 34.94 & 32.71 & 51.31 & 33.20 & 57.50 & \textbf{31.70} \\
NextExcl$\uparrow$ & 0.00 & 3.67 & 71.34 & 69.47 & 50.50 & 68.46 & 44.46 & \textbf{74.05} \\
PayloadHit$\downarrow$ & 48.07 & 40.42 & 25.90 & 36.30 & 3.38 & 27.49 & 19.85 & \textbf{0.05} \\
CF-Harm$\downarrow$ & 36.06 & 29.34 & 7.52 & 9.84 & \textbf{2.23} & 7.43 & 11.33 & 2.78 \\
\bottomrule
\end{tabularx}
\end{table}

\begin{table}[!htb]
\centering
\caption{Strict wins for each agent composition. Bold marks the largest count in each row. Shaded cells are summarized in panel (a) of Table~\ref{tab:main_result_summary}.}
\label{tab:configuration_wins}
\footnotesize
\setlength{\tabcolsep}{3pt}
\begin{tabularx}{\linewidth}{@{}>{\raggedright\arraybackslash}p{0.22\linewidth}*{8}{>{\raggedleft\arraybackslash}X}@{}}
\toprule
Metric / setting & UMaj & URand & Single & Babylon & Eigen & Beta & TS & \rep \\
\midrule
\multicolumn{9}{@{}l}{\textit{GoEmotions / exact-match accuracy}} \\
All strong & 0/28 & 0/28 & 0/28 & 0/28 & 0/28 & 6/28 & 0/28 & \textbf{15/28} \\
7 strong + 3 weak & \cellcolor{blue!10}0/28 & \cellcolor{blue!10}0/28 & \cellcolor{blue!10}0/28 & \cellcolor{blue!10}0/28 & \cellcolor{blue!10}0/28 & \cellcolor{blue!10}0/28 & \cellcolor{blue!10}0/28 & \cellcolor{blue!10}\textbf{25/28} \\
All weak & 0/28 & 0/28 & 2/28 & 0/28 & 0/28 & 2/28 & 0/28 & \textbf{18/28} \\
Random & 0/28 & 0/28 & 1/28 & 3/28 & 0/28 & \textbf{9/28} & 0/28 & 6/28 \\
\midrule
\multicolumn{9}{@{}l}{\textit{GoEmotions / sample F1}} \\
All strong & 0/28 & 0/28 & 1/28 & 3/28 & 0/28 & \textbf{11/28} & 1/28 & 9/28 \\
7 strong + 3 weak & 0/28 & 0/28 & 2/28 & 0/28 & 0/28 & \textbf{13/28} & 7/28 & 3/28 \\
All weak & 0/28 & 0/28 & \cellcolor{blue!10}8/28 & 0/28 & 0/28 & 3/28 & 0/28 & \cellcolor{blue!10}\textbf{17/28} \\
Random & 0/28 & 0/28 & \textbf{7/28} & 1/28 & 0/28 & \textbf{7/28} & 6/28 & 5/28 \\
\midrule
\multicolumn{9}{@{}l}{\textit{MATH / mathematical-equivalence accuracy}} \\
All strong & \cellcolor{blue!10}0/28 & \cellcolor{blue!10}0/28 & \cellcolor{blue!10}0/28 & \cellcolor{blue!10}0/28 & \cellcolor{blue!10}0/28 & \cellcolor{blue!10}0/28 & \cellcolor{blue!10}0/28 & \cellcolor{blue!10}\textbf{27/28} \\
7 strong + 3 weak & \cellcolor{blue!10}0/28 & \cellcolor{blue!10}0/28 & \cellcolor{blue!10}0/28 & \cellcolor{blue!10}0/28 & \cellcolor{blue!10}0/28 & \cellcolor{blue!10}0/28 & \cellcolor{blue!10}0/28 & \cellcolor{blue!10}\textbf{28/28} \\
All weak & 0/28 & 0/28 & 2/28 & 0/28 & 0/28 & 7/28 & 1/28 & \textbf{10/28} \\
Random & 0/28 & 0/28 & 0/28 & 0/28 & 0/28 & 2/28 & 1/28 & \textbf{19/28} \\
\midrule
\multicolumn{9}{@{}l}{\textit{HumanEval Pro / Pass@1}} \\
All strong & 0/28 & 0/28 & 2/28 & 1/28 & \textbf{9/28} & 0/28 & 0/28 & 4/28 \\
7 strong + 3 weak & 0/28 & 0/28 & 1/28 & 0/28 & \cellcolor{blue!10}10/28 & 0/28 & 1/28 & \cellcolor{blue!10}\textbf{12/28} \\
All weak & 0/28 & 1/28 & 2/28 & 1/28 & 6/28 & 0/28 & 1/28 & \textbf{11/28} \\
Random & 0/28 & 0/28 & 2/28 & 1/28 & 6/28 & 0/28 & 2/28 & \textbf{9/28} \\
\bottomrule
\end{tabularx}
\end{table}

\begin{table}[!htb]
\centering
\caption{Paired stress cases in the 7S+3W composition. Baseline $b$ has the highest attacked score, where ties are resolved as mentioned in \appref{app:average_protocol}. Scores and CF-Harm are percentages; $\Delta$ and $R_b$ are percentage points. Shaded cells are reported in \tabref{tab:main_result_summary}(b). Bold marks the best value in each row.}
\label{tab:paired_preservation}
\footnotesize
\setlength{\tabcolsep}{3pt}
\begin{tabularx}{\linewidth}{@{}>{\raggedright\arraybackslash}p{0.22\linewidth}*{8}{>{\raggedleft\arraybackslash}X}@{}}
\toprule
Metric / setting & UMaj & URand & Single & Babylon & Eigen & Beta & TS & \rep \\
\midrule
\multicolumn{9}{@{}l}{\textit{GoEmotions / B / \textup{\textsc{Strong}} (F1; $b=$ Beta; $R_b=+0.4$)}} \\
Clean$\uparrow$ & 37.40 & 34.50 & 37.37 & 37.10 & 36.10 & 38.27 & \textbf{38.43} & 35.00 \\
Attacked$\uparrow$ & 18.63 & 24.33 & 35.27 & 32.30 & 21.90 & \cellcolor{blue!10}\textbf{35.83} & 35.33 & \cellcolor{blue!10}33.00 \\
$\Delta\downarrow$ & 18.77 & 10.17 & 2.10 & 4.80 & 14.20 & 2.43 & 3.10 & \textbf{2.00} \\
CF-Harm$\downarrow$ & 48.28 & 31.03 & 6.90 & 12.64 & 36.78 & 8.05 & 9.20 & \textbf{3.45} \\
\midrule
\multicolumn{9}{@{}l}{\textit{GoEmotions / B / \textup{\textsc{Arith}} (F1; $b=$ Beta; $R_b=+4.2$)}} \\
Attacked$\uparrow$ & 28.87 & 28.20 & 29.60 & 30.50 & 29.17 & \cellcolor{blue!10}34.00 & 27.83 & \cellcolor{blue!10}\textbf{34.90} \\
$\Delta\downarrow$ & 8.53 & 6.30 & 7.77 & 6.60 & 6.93 & 4.27 & 10.60 & \textbf{0.10} \\
CF-Harm$\downarrow$ & 32.91 & 24.05 & 22.78 & 21.52 & 24.05 & 16.46 & 32.91 & \textbf{5.06} \\
\midrule
\multicolumn{9}{@{}l}{\textit{GoEmotions / B / \textup{\textsc{Bound}} (F1; $b=$ Single; $R_b=+3.8$)}} \\
Attacked$\uparrow$ & 26.03 & 25.97 & \cellcolor{blue!10}33.73 & 30.60 & 25.93 & 31.57 & 27.27 & \cellcolor{blue!10}\textbf{35.17} \\
$\Delta\downarrow$ & 11.37 & 8.53 & 3.63 & 6.50 & 10.17 & 6.70 & 11.17 & \textbf{-0.17} \\
CF-Harm$\downarrow$ & 38.27 & 29.63 & 18.52 & 22.22 & 29.63 & 20.99 & 32.10 & \textbf{3.70} \\
\midrule
\multicolumn{9}{@{}l}{\textit{MATH / R2 / \textup{\textsc{Div}} (accuracy; $b=$ Babylon; $R_b=+4.0$)}} \\
Clean$\uparrow$ & 73.00 & 67.00 & 66.00 & 65.00 & 67.00 & 71.00 & 65.00 & \textbf{79.00} \\
Attacked$\uparrow$ & 53.00 & 48.00 & 53.00 & \cellcolor{blue!10}55.00 & 47.00 & 54.00 & 54.00 & \cellcolor{blue!10}\textbf{73.00} \\
$\Delta\downarrow$ & 20.00 & 19.00 & 13.00 & 10.00 & 20.00 & 17.00 & 11.00 & \textbf{6.00} \\
CF-Harm$\downarrow$ & 26.32 & 25.00 & 15.79 & 17.11 & 26.32 & 21.05 & 14.47 & \textbf{11.84} \\
\midrule
\multicolumn{9}{@{}l}{\textit{MATH / R2 / \textup{\textsc{Bound}} (accuracy; $b=$ Babylon; $R_b=+1.0$)}} \\
Attacked$\uparrow$ & 37.00 & 38.00 & 53.00 & \cellcolor{blue!10}54.00 & 37.00 & 53.00 & 48.00 & \cellcolor{blue!10}\textbf{69.00} \\
$\Delta\downarrow$ & 36.00 & 29.00 & 13.00 & 11.00 & 30.00 & 18.00 & 17.00 & \textbf{10.00} \\
CF-Harm$\downarrow$ & 43.37 & 34.94 & \textbf{16.87} & 18.07 & 37.35 & 22.89 & 20.48 & \textbf{16.87} \\
\bottomrule
\end{tabularx}
\end{table}

\begin{table}[!htb]
\centering
\caption{Performance under OnOff and Adapt attacks on HumanEval Pro under the 7S+3W composition. Baseline $b$ has the highest attacked Pass@1 in each setting, where ties are resolved as mentioned in \appref{app:average_protocol}. Shaded cells are reported in \tabref{tab:main_result_summary}(c). ``--'' denotes undefined PayloadHit. Bold marks the best defined value in each row.}
\label{tab:late_attack_code}
\footnotesize
\setlength{\tabcolsep}{3pt}
\begin{tabularx}{\linewidth}{@{}>{\raggedright\arraybackslash}p{0.22\linewidth}*{8}{>{\raggedleft\arraybackslash}X}@{}}
\toprule
Metric / setting & UMaj & URand & Single & Babylon & Eigen & Beta & TS & \rep \\
\midrule
\multicolumn{9}{@{}l}{\textit{HumanEval Pro / Pass@1 (shared clean reference)}} \\
Clean$\uparrow$ & \textbf{83.00} & 80.00 & 79.00 & 78.00 & 79.00 & 79.00 & 77.00 & 79.00 \\
\midrule
\multicolumn{9}{@{}l}{\textit{HumanEval Pro / B / \textup{\textsc{OnOff}} ($b=$ Eigen)}} \\
Attacked$\uparrow$ & 45.00 & 51.00 & 77.00 & 75.00 & \cellcolor{blue!10}78.00 & 75.00 & 51.00 & \cellcolor{blue!10}\textbf{81.00} \\
CF-Harm$\downarrow$ & 54.29 & 41.43 & 5.71 & 7.14 & 4.29 & 8.57 & 40.00 & \textbf{1.43} \\
PayloadHit$\downarrow$ & 64.29 & 52.94 & 16.67 & 44.44 & 14.29 & 35.71 & 54.29 & \textbf{0.00} \\
\midrule
\multicolumn{9}{@{}l}{\textit{HumanEval Pro / B / \textup{\textsc{Adapt}} ($b=$ UMaj)}} \\
Attacked$\uparrow$ & \cellcolor{blue!10}\textbf{81.00} & 78.00 & 80.00 & 79.00 & 79.00 & 78.00 & 80.00 & \cellcolor{blue!10}\textbf{81.00} \\
CF-Harm$\downarrow$ & 4.08 & 4.08 & 4.08 & 6.12 & 6.12 & 8.16 & \textbf{2.04} & \textbf{2.04} \\
PayloadHit$\downarrow$ & 10.20 & 10.81 & 7.14 & 10.00 & 7.89 & 7.89 & 2.94 & \textbf{0.00} \\
\midrule
\multicolumn{9}{@{}l}{\textit{HumanEval Pro / R1 / \textup{\textsc{OnOff}} ($b=$ Eigen)}} \\
Attacked$\uparrow$ & 46.00 & 50.00 & 77.00 & 75.00 & \cellcolor{blue!10}81.00 & 76.00 & 64.00 & \cellcolor{blue!10}\textbf{82.00} \\
CF-Harm$\downarrow$ & 52.86 & 42.86 & 5.71 & 8.57 & \textbf{0.00} & 7.14 & 22.86 & \textbf{0.00} \\
PayloadHit$\downarrow$ & 64.29 & 52.86 & 20.00 & 66.67 & \textbf{0.00} & 62.50 & 28.57 & \textbf{0.00} \\
\midrule
\multicolumn{9}{@{}l}{\textit{HumanEval Pro / R1 / \textup{\textsc{Adapt}} ($b=$ Eigen)}} \\
Attacked$\uparrow$ & 78.00 & 75.00 & 72.00 & 78.00 & \cellcolor{blue!10}\textbf{82.00} & 78.00 & 78.00 & \cellcolor{blue!10}79.00 \\
CF-Harm$\downarrow$ & 13.51 & 13.51 & 18.92 & 5.41 & \textbf{0.00} & 8.11 & 5.41 & 2.70 \\
PayloadHit$\downarrow$ & 8.11 & 23.08 & 13.79 & 8.33 & \textbf{0.00} & 8.57 & 6.25 & \textbf{0.00} \\
\midrule
\multicolumn{9}{@{}l}{\textit{HumanEval Pro / R2 / \textup{\textsc{OnOff}} ($b=$ Eigen)}} \\
Attacked$\uparrow$ & 45.00 & 47.00 & 78.00 & 75.00 & \cellcolor{blue!10}79.00 & 75.00 & 56.00 & \cellcolor{blue!10}\textbf{82.00} \\
CF-Harm$\downarrow$ & 54.29 & 47.14 & 4.29 & 7.14 & 2.86 & 8.57 & 31.43 & \textbf{1.43} \\
PayloadHit$\downarrow$ & 64.29 & 58.57 & 20.00 & 50.00 & 8.33 & 41.67 & 50.00 & \textbf{0.00} \\
\midrule
\multicolumn{9}{@{}l}{\textit{HumanEval Pro / R2 / \textup{\textsc{Adapt}} ($b=$ Eigen)}} \\
Attacked$\uparrow$ & 80.00 & 77.00 & 78.00 & 77.00 & \cellcolor{blue!10}80.00 & 78.00 & 77.00 & \cellcolor{blue!10}\textbf{83.00} \\
CF-Harm$\downarrow$ & 7.32 & 7.32 & 9.76 & 12.20 & 2.44 & 9.76 & 7.32 & \textbf{0.00} \\
PayloadHit$\downarrow$ & 12.20 & 15.38 & 10.00 & 10.81 & 5.13 & 12.82 & 14.29 & \textbf{0.00} \\
\midrule
\multicolumn{9}{@{}l}{\textit{HumanEval Pro / W / \textup{\textsc{OnOff}} ($b=$ Babylon)}} \\
Attacked$\uparrow$ & 48.00 & 57.00 & 80.00 & \cellcolor{blue!10}80.00 & 79.00 & 80.00 & 79.00 & \cellcolor{blue!10}\textbf{81.00} \\
CF-Harm$\downarrow$ & 50.00 & 32.86 & 1.43 & \textbf{0.00} & 1.43 & 1.43 & \textbf{0.00} & \textbf{0.00} \\
PayloadHit$\downarrow$ & 61.43 & 42.86 & \textbf{0.00} & \textbf{0.00} & \textbf{0.00} & \textbf{0.00} & \textbf{0.00} & -- \\
\midrule
\multicolumn{9}{@{}l}{\textit{HumanEval Pro / W / \textup{\textsc{Adapt}} ($b=$ Babylon)}} \\
Attacked$\uparrow$ & \textbf{81.00} & 79.00 & \textbf{81.00} & \cellcolor{blue!10}\textbf{81.00} & 80.00 & 80.00 & 79.00 & \cellcolor{blue!10}\textbf{81.00} \\
CF-Harm$\downarrow$ & 6.45 & 3.23 & \textbf{0.00} & \textbf{0.00} & 3.23 & 3.23 & \textbf{0.00} & \textbf{0.00} \\
PayloadHit$\downarrow$ & 9.68 & 9.09 & \textbf{0.00} & \textbf{0.00} & \textbf{0.00} & \textbf{0.00} & \textbf{0.00} & \textbf{0.00} \\
\bottomrule
\end{tabularx}
\end{table}

\heading{Performance under different compositions, expanded.} \tabref{tab:configuration_wins} expands panel (a) of \tabref{tab:main_result_summary} by reporting the strict wins of all methods under all four agent compositions. The results confirm that \rep performs most consistently on MATH: it achieves 27, 28, 10, and 19 strict wins under the all-strong, 7S+3W, all-weak, and random compositions, respectively, for a total of 84 wins across the 112 conditions. Its performance on the other datasets depends more strongly on the composition. On GoEmotions, \rep performs particularly well in exact-match accuracy under the 7S+3W composition and in F1 under the all-weak composition, but Beta or Single-Metric obtains more strict wins in several other settings. On HumanEval Pro, \rep achieves the most strict wins under the 7S+3W, all-weak, and random compositions, while EigenTrust performs better under the all-strong composition. We believe these differences reflect how the available proposal quality changes with the agent composition. Reputation-based aggregation is most useful when the pool contains meaningful differences in agent capability. Its advantage can decrease when another method already assigns high influence to the strongest proposals.

\heading{Performance when capable agents are compromised.} \tabref{tab:paired_preservation} expands \tabref{tab:main_result_summary}(b) with the clean score, attacked score, paired loss, and CF-Harm for five cases in the 7S+3W composition. The compromised agents include highly ranked models: the GoEmotions/B setting compromises Qwen-Max, Qwen-Flash, and GLM, while the MATH/R2 setting compromises two Kimi agents and one GLM agent. \rep achieves the highest attacked score in four of the five cases and has a positive $R_b$ in all five. On GoEmotions, it exceeds the strongest baseline under Arith and Bound, with paired losses of only 0.10 and $-0.17$ points. Under Strong, its attacked F1 remains below Beta, although it has a smaller paired loss and lower CF-Harm. On MATH, \rep achieves 73\% and 69\% accuracy under Div and Bound, exceeding Babylon by 18 and 15 percentage points, respectively. These results are consistent with \rep using the current proposal to adjust an agent's influence, so a capable agent cannot rely only on its historical reputation after it begins submitting malicious proposals.

\heading{Performance under OnOff and Adapt attacks, expanded.} \tabref{tab:late_attack_code} expands \tabref{tab:main_result_summary}(c) by reporting Pass@1, CF-Harm, and PayloadHit for the OnOff and Adapt attacks under all four corruption placements on HumanEval Pro. \rep has five strict wins, two ties, and one loss against the strongest baseline in these eight conditions. It strictly outperforms the strongest baseline under all four OnOff placements. Under Adapt, it achieves one strict win and two ties. Among them, R1 is the only exception, where \rep obtains 79\% Pass@1 and EigenTrust obtains 82\%. When PayloadHit is defined, the final output never matches an injected answer for \rep in any of these conditions. Under W/OnOff, PayloadHit is undefined because no proposal containing an attack payload enters the aggregation pool. These results are consistent with \rep considering both the current proposal and historical reputation, which limits the influence of agents that behave honestly before launching an attack or change which agents actively attack.

\heading{Evaluation scope.} \label{app:evaluation-scope}
These fixed-task results assume persistent identities and known clone groups. The 112 conditions are not independent task samples or generation-seed repetitions. Beyond the two held-out families examined in \appref{app:feedback-generalization}, transfer to additional attack families, multi-round debate, and white-box attacks against the screening models remains untested here.

\clearpage
\subsection{Generalization to Unseen Attack Types}
\label{app:feedback-generalization}
\heading{Experimental setup.} We evaluate \rep on attack families excluded from feedback-model training to assess whether its task performance extends beyond the corruption patterns seen during training.
We reuse the main experiments' fixed tasks, four agent compositions, four corruption placements, and eight aggregation methods.
Within each composition--placement setting, we vary the test attack family while keeping the aggregation configuration, feedback-model checkpoints, and calibrated thresholds unchanged, without retraining.
\tabref{tab:heldout_attack_scores} compares test performance on the five families represented in training (\textsc{Rand}, \textsc{Strong}, \textsc{Div}, \textsc{OnOff}, and \textsc{Adapt}) with the held-out \textsc{Arith} and \textsc{Bound} families.
We retain the main task metrics: GoEmotions exact-match accuracy and sample F1, MATH mathematical-equivalence accuracy, and HumanEval Pro Pass@1.

\begin{table}[!htbp]
\centering
\caption{Test performance on the five attack families represented in feedback-model training and the two held-out families. First five averages $4\times4\times5=80$ conditions; each held-out family averages $4\times4=16$. Scores are percentages; bold marks the largest mean in each row. These are complete-system comparisons, not feedback-removal ablations.}
\label{tab:heldout_attack_scores}
\footnotesize
\setlength{\tabcolsep}{3pt}
{\begin{tabularx}{\linewidth}{@{}>{\raggedright\arraybackslash}p{0.22\linewidth}*{8}{>{\raggedleft\arraybackslash}X}@{}}
\toprule
Metric / setting & UMaj & URand & Single & Babylon & Eigen & Beta & TS & \rep \\
\midrule
\multicolumn{9}{@{}l}{\textit{GoEmotions / exact-match accuracy}} \\
First five & 11.59 & 11.59 & 19.55 & 18.54 & 12.91 & 19.76 & 19.00 & \textbf{21.33} \\
\textsc{Arith} & 9.00 & 9.50 & 18.44 & 17.31 & 12.31 & 20.00 & 15.06 & \textbf{20.94} \\
\textsc{Bound} & 8.00 & 9.81 & 17.06 & 16.38 & 10.06 & 18.31 & 14.44 & \textbf{20.56} \\
\midrule
\multicolumn{9}{@{}l}{\textit{GoEmotions / sample F1}} \\
First five & 28.04 & 27.82 & 34.72 & 34.09 & 28.44 & \textbf{35.06} & 34.71 & 33.95 \\
\textsc{Arith} & 28.33 & 27.32 & 33.17 & 33.29 & 29.79 & \textbf{35.21} & 31.32 & 33.97 \\
\textsc{Bound} & 26.71 & 27.89 & 32.21 & 32.34 & 27.70 & 33.58 & 30.39 & \textbf{33.87} \\
\midrule
\multicolumn{9}{@{}l}{\textit{MATH / mathematical-equivalence accuracy}} \\
First five & 45.83 & 45.13 & 54.06 & 53.03 & 43.70 & 54.51 & 49.01 & \textbf{62.05} \\
\textsc{Arith} & 39.94 & 40.63 & 53.50 & 51.94 & 38.75 & 53.69 & 47.44 & \textbf{61.00} \\
\textsc{Bound} & 37.19 & 38.13 & 53.13 & 50.69 & 36.81 & 54.31 & 47.44 & \textbf{62.38} \\
\midrule
\multicolumn{9}{@{}l}{\textit{HumanEval Pro / Pass@1}} \\
First five & 54.79 & 58.41 & 75.63 & 73.61 & 77.84 & 74.75 & 72.16 & \textbf{78.19} \\
\textsc{Arith} & 43.00 & 49.25 & 73.00 & 69.88 & 78.06 & 72.38 & 67.56 & \textbf{78.25} \\
\textsc{Bound} & 45.31 & 51.63 & 71.69 & 67.44 & \textbf{78.19} & 70.94 & 68.69 & 77.88 \\
\bottomrule
\end{tabularx}}
\end{table}

\heading{Results.} On MATH, \rep reaches 61.00\% and 62.38\% accuracy under \textsc{Arith} and \textsc{Bound}, exceeding the best baseline means by 7.31 and 8.06 percentage points.
Its GoEmotions exact-match accuracy is also highest on both held-out families, whereas sample F1 is mixed; HumanEval Pro Pass@1 differs from EigenTrust by only $+0.19$ and $-0.31$ points, respectively.
Thus the complete system retains useful performance on attack families excluded from feedback-model training, without uniformly improving every task metric.

\newpage
\heading{Feedback diagnostics.} We also examine how the learned feedback responds to held-out attacks, since task scores alone do not distinguish proposal-quality estimation from attack identification.
Using the same runs and fixed checkpoints as above, we pool the recorded proposal-level counts within each dataset and attack group.
\tabref{tab:heldout_feedback_diagnostics} reports semantic low-quality prediction rates for proposals with and without active payloads, together with recall and false-positive rates for the behavior probe and fused detector.

\begin{table}[!htbp]
\centering
\caption{Recorded proposal-level feedback diagnostics (\%). Counts are pooled within each dataset and attack group before computing rates; each configuration is counted once. P and N denote proposals with and without an active attack payload. Probe and fused-detector recall/FPR use the active-payload label. The semantic columns measure low-quality predictions: N can include naturally incorrect proposals, so its low-quality rate is not a false-positive rate.}
\label{tab:heldout_feedback_diagnostics}
\footnotesize
\setlength{\tabcolsep}{4pt}
{\begin{tabularx}{\linewidth}{@{}>{\raggedright\arraybackslash}p{0.22\linewidth}*{6}{>{\raggedleft\arraybackslash}X}@{}}
\toprule
& \multicolumn{2}{c}{Semantic low-quality} & \multicolumn{2}{c}{Behavior probe} & \multicolumn{2}{c}{Fused detector} \\
\cmidrule(lr){2-3}\cmidrule(lr){4-5}\cmidrule(lr){6-7}
Dataset / attack & P & N & Recall & FPR & Recall & FPR \\
\midrule
\multicolumn{7}{@{}l}{\textit{GoEmotions}} \\
First five & 91.40 & 60.03 & 23.55 & 9.69 & 24.77 & 12.30 \\
\textsc{Arith} & 78.51 & 59.39 & 21.73 & 9.58 & 25.18 & 12.92 \\
\textsc{Bound} & 80.61 & 57.99 & 21.68 & 9.74 & 26.34 & 13.05 \\
\midrule
\multicolumn{7}{@{}l}{\textit{MATH}} \\
First five & 98.32 & 45.21 & 29.94 & 1.24 & 34.75 & 9.42 \\
\textsc{Arith} & 97.99 & 44.71 & 32.25 & 1.21 & 35.60 & 9.82 \\
\textsc{Bound} & 97.94 & 43.48 & 31.40 & 1.22 & 34.64 & 9.20 \\
\midrule
\multicolumn{7}{@{}l}{\textit{HumanEval Pro}} \\
First five & 31.47 & 6.32 & 14.43 & 1.44 & 9.12 & 4.98 \\
\textsc{Arith} & 27.94 & 6.19 & 17.94 & 1.47 & 10.44 & 4.79 \\
\textsc{Bound} & 31.40 & 6.79 & 17.47 & 1.31 & 10.44 & 4.99 \\
\bottomrule
\end{tabularx}}
\end{table}

\heading{Results.} For MATH, the semantic verifier assigns low quality to 97.99\% and 97.94\% of payload proposals under \textsc{Arith} and \textsc{Bound}, while fused-detector recall is 35.60\% and 34.64\%.
For HumanEval Pro, fused recall is only about 10.44\% on each held-out family despite end-to-end Pass@1 near 78\%.
These diagnostics distinguish proposal-quality screening from attack identification; neither they nor the system scores isolate the incremental benefit of learned feedback, which requires a matched feedback-removal ablation.

\clearpage
\subsection{Performance under Different Agent Compositions}
\label{app:composition-averages}
\heading{Experimental setup.} We test how agent composition affects task performance and the relative ranking of aggregation methods, to determine whether the main results depend on a particular mix of strong and weak models.
We reuse the main $4\times4\times7$ experiment grid and vary its first factor across all-strong, 7S+3W, all-weak, and random ten-agent pools.
The fixed tasks, method configurations, learned-feedback settings, and task metrics remain as in the main experiments.
Each composition is evaluated under the same four corruption-placement rules and seven attack families; all methods share cached proposals within each condition.
We average task scores over these 28 conditions for each composition, so \tabref{tab:composition_average_scores} shows the magnitude of performance differences alongside the strict-win counts in \tabref{tab:configuration_wins}.

\begin{table}[!htbp]
\centering
\caption{Mean attacked task scores (\%) under all four agent compositions. Every entry averages the four corruption placements and seven attacks ($28$ conditions), with the method configuration held fixed. Bold marks the largest mean in each row. Unlike the strict-win counts in \tabref{tab:configuration_wins}, this table retains score magnitudes.}
\label{tab:composition_average_scores}
\footnotesize
\setlength{\tabcolsep}{3pt}
{\begin{tabularx}{\linewidth}{@{}>{\raggedright\arraybackslash}p{0.22\linewidth}*{8}{>{\raggedleft\arraybackslash}X}@{}}
\toprule
Metric / setting & UMaj & URand & Single & Babylon & Eigen & Beta & TS & \rep \\
\midrule
\multicolumn{9}{@{}l}{\textit{GoEmotions / exact-match accuracy}} \\
All strong & 13.14 & 14.18 & 21.46 & 21.46 & 14.64 & 22.71 & 21.21 & \textbf{23.79} \\
7 strong + 3 weak & 10.32 & 10.00 & 19.71 & 17.82 & 11.89 & 19.93 & 18.54 & \textbf{23.50} \\
All weak & 6.89 & 7.89 & 14.07 & 11.93 & 8.82 & 13.61 & 11.61 & \textbf{15.82} \\
Random & 12.46 & 12.07 & 20.89 & 21.00 & 14.32 & \textbf{22.11} & 19.79 & 21.54 \\
\midrule
\multicolumn{9}{@{}l}{\textit{GoEmotions / sample F1}} \\
All strong & 29.50 & 29.71 & 34.70 & 35.46 & 29.86 & \textbf{36.51} & 35.05 & 35.37 \\
7 strong + 3 weak & 28.79 & 28.22 & 35.93 & 35.08 & 28.92 & \textbf{36.80} & 35.73 & 34.73 \\
All weak & 24.47 & 24.47 & 30.28 & 28.60 & 25.56 & 29.87 & 28.58 & \textbf{31.65} \\
Random & 28.80 & 28.62 & 35.65 & 35.75 & 29.77 & \textbf{36.31} & 35.10 & 34.00 \\
\midrule
\multicolumn{9}{@{}l}{\textit{MATH / mathematical-equivalence accuracy}} \\
All strong & 58.50 & 56.93 & 71.75 & 70.61 & 57.89 & 71.71 & 63.61 & \textbf{78.82} \\
7 strong + 3 weak & 51.14 & 48.54 & 64.89 & 62.36 & 47.75 & 65.75 & 55.61 & \textbf{76.57} \\
All weak & 28.61 & 31.96 & 35.57 & 34.21 & 27.36 & \textbf{36.04} & 34.11 & 35.71 \\
Random & 36.75 & 36.50 & 43.18 & 42.96 & 35.04 & 43.96 & 40.93 & \textbf{56.68} \\
\midrule
\multicolumn{9}{@{}l}{\textit{HumanEval Pro / Pass@1}} \\
All strong & 53.50 & 59.54 & 77.11 & 74.54 & \textbf{79.93} & 77.11 & 76.36 & 78.82 \\
7 strong + 3 weak & 53.50 & 56.18 & 75.39 & 72.93 & 79.43 & 74.39 & 71.11 & \textbf{79.68} \\
All weak & 50.43 & 55.50 & 71.68 & 68.96 & 74.79 & 70.68 & 66.29 & \textbf{76.11} \\
Random & 49.57 & 53.32 & 74.57 & 72.36 & 77.54 & 73.29 & 70.29 & \textbf{78.00} \\
\bottomrule
\end{tabularx}}
\end{table}

\heading{Results.} On MATH, \rep exceeds the best baseline mean by 7.07, 10.82, and 12.71 percentage points in the all-strong, 7S+3W, and random compositions, respectively.
The all-weak composition is the exception: \rep scores 35.71\%, slightly below Beta's 36.04\%, even though it has the strictest wins (10 versus 7).
This difference shows why a larger win count need not imply a larger mean score.

Composition also changes the relative rankings on the other tasks.
On GoEmotions, \rep leads in exact-match accuracy for all-strong, 7S+3W, and all-weak pools, but Beta leads for random pools; for sample F1, \rep leads only in the all-weak composition.
On HumanEval Pro, EigenTrust leads in the all-strong composition, whereas \rep has the largest mean in the other three, with margins of 0.25--1.32 points.
These comparisons measure sensitivity to the evaluated pool compositions rather than isolate any one reputation component or establish significance across independent generation seeds.

\end{document}